\documentclass[10pt,journal,compsoc]{IEEEtran}

\ifCLASSOPTIONcompsoc
  \usepackage[nocompress]{cite}
\else
  \usepackage{cite}
\fi

\ifCLASSINFOpdf

\else

\fi

\usepackage{amssymb}
\usepackage{cite}
\usepackage{graphicx}
\usepackage{subfigure}
\usepackage{bbding}
\usepackage{multirow}
\usepackage{amsmath}
\usepackage{amssymb}
\usepackage{epstopdf}
\usepackage{url}
\usepackage{hyperref}
\usepackage{amsthm}

\usepackage{booktabs}
\usepackage{bm}
\usepackage{cite}
\usepackage{fancyhdr}
\usepackage{amscd}
\usepackage{amsfonts}
\usepackage{algorithm}
\usepackage{algorithmic}
\usepackage{chngpage}
\usepackage{multirow}
\usepackage[table,xcdraw]{xcolor}
\usepackage{fancyhdr}
\usepackage{amscd}
\usepackage[justification=centering]{caption}
\usepackage{booktabs}  
\usepackage{threeparttable}  
\usepackage{makecell}
\usepackage{graphicx}
\usepackage{array}
\usepackage{textcomp}
\usepackage{xcolor}
\usepackage{booktabs}
\usepackage{mathrsfs}
\usepackage{url}
\usepackage{amsthm,amssymb,mathtools}
\usepackage{mathrsfs}
\usepackage{booktabs}

\usepackage{float}
\newtheorem{definition}{Definition}
\usepackage{xspace}

\begin{document}

\newcommand{\method}{AMSL\xspace}
\newcommand{\methodfull}{Adaptive Memory Network with Self-supervised Learning\xspace}
\newcommand{\uad}{unsupervised anomaly detection\xspace}
%
\title{Adaptive Memory Networks with Self-supervised Learning for Unsupervised Anomaly Detection}
%
%
%
%

\author{Yuxin~Zhang,
        Jindong~Wang,
        Yiqiang~Chen,
        Han~Yu,
        Tao~Qin

\IEEEcompsocitemizethanks{
\IEEEcompsocthanksitem Y. Zhang is with Global Energy Interconnection Development and Cooperation Organization, Xicheng District, Beijing, China, and Beijing Key Laboratory of Mobile Computing and Pervasive Device, Institute of Computing Technology, Chinese Academy of Sciences and University of Chinese Academy of Sciences, Beijing, China.
E-mail: yuxin-zhang@geidco.org.

\IEEEcompsocthanksitem Y. Chen are with Beijing Key Laboratory of Mobile Computing and Pervasive Device, Institute of Computing Technology, Chinese Academy of Sciences, Beijing, China and University of Chinese Academy of Sciences, Beijing, China. Y. Chen is also with Peng cheng Laboratory (PCL).
E-mail: yqchen@ict.ac.cn.
\IEEEcompsocthanksitem J. Wang and T. Qin are with Microsoft Research, Beijing, China.
	E-mail: \{jindong.wang, taoqin\}@microsoft.com.

\IEEEcompsocthanksitem H. Yu is with the School of Computer Science and Engineering, Nanyang Technological University, Singapore.
	E-mail: han.yu@ntu.egu.sg.
	
\IEEEcompsocthanksitem Corresponding author: Yiqiang Chen and Jindong Wang}

\thanks{Manuscript received April 19, 2005; revised August 26, 2015.}}

%
%

\markboth{}%
{Shell \MakeLowercase{\textit{et al.}}: Bare Demo of IEEEtran.cls for Computer Society Journals}
%



\IEEEtitleabstractindextext{%
\begin{abstract}
Unsupervised anomaly detection aims to build models to effectively detect unseen anomalies by only training on the normal data. Although previous reconstruction-based methods have made fruitful progress, their generalization ability is limited due to two critical challenges. First, the training dataset only contains normal patterns, which limits the model generalization ability. Second, the feature representations learned by existing models often lack representativeness which hampers the ability to preserve the diversity of normal patterns. In this paper, we propose a novel approach called Adaptive Memory Network with Self-supervised Learning (AMSL) to address these challenges and enhance the generalization ability in unsupervised anomaly detection. Based on the convolutional autoencoder structure, AMSL incorporates a self-supervised learning module to learn general normal patterns and an adaptive memory fusion module to learn rich feature representations. Experiments on four public multivariate time series datasets demonstrate that AMSL significantly improves the performance compared to other state-of-the-art methods. Specifically, on the largest CAP sleep stage detection dataset with 900 million samples, AMSL outperforms the second-best baseline by \textbf{4}\%+ in both accuracy and F1 score. 
Apart from the enhanced generalization ability, AMSL is also more robust against input noise.
\end{abstract}

}

\maketitle

\IEEEdisplaynontitleabstractindextext

%
\IEEEpeerreviewmaketitle

\IEEEraisesectionheading{\section{Introduction}}
\label{sec-intro}

\IEEEPARstart{W}{ith} the prevalence of Internet of Things (IoT) devices in many applications (e.g., health care, human activity recognition and industrial control systems), there is growing interest in anomaly detection \cite{kiran2018overview,mirsky2018kitsune,chuah2007ecg}.
However, it is prohibitively expensive to acquire large amounts of labeled anomaly data required to train machine learning models following the current approaches.
For instance, the collection of sleep apnea data \cite{terzano2002atlas} is extremely time-consuming and restricted by the environment, while the data for the normal states are much easier to obtain.
For such scenarios, \uad is a promising paradigm that aims to detect anomalies by training from only normal samples, which is our main focus in this paper, and specifically, for multivariate time series.

The key challenge to \uad is to learn generalized patterns from the normal training data such that it can achieve good performance on the \emph{unseen} anomalies.
Over the years, there have been a line of works on this topic.
Autoencoder (AE) is a powerful unsupervised learning technique that has been widely used for unsupervised anomaly detection \cite{sakurada2014anomaly,gutoski2017detection,hasan2016learning}.
AE is usually trained by minimizing the reconstruction error on the normal data, and then using such errors as indicators or thresholds of anomalies.
Based on the AE structure, LSTM-AE \cite{malhotra2016lstm}, Convolutional AE (CAE) \cite{gutoski2017detection}, and ConvLSTM-AE \cite{zhang2019deep} have emerged as promising anomaly detection approaches.

\begin{figure}[t!]
\centering
\includegraphics[width=0.48\textwidth]{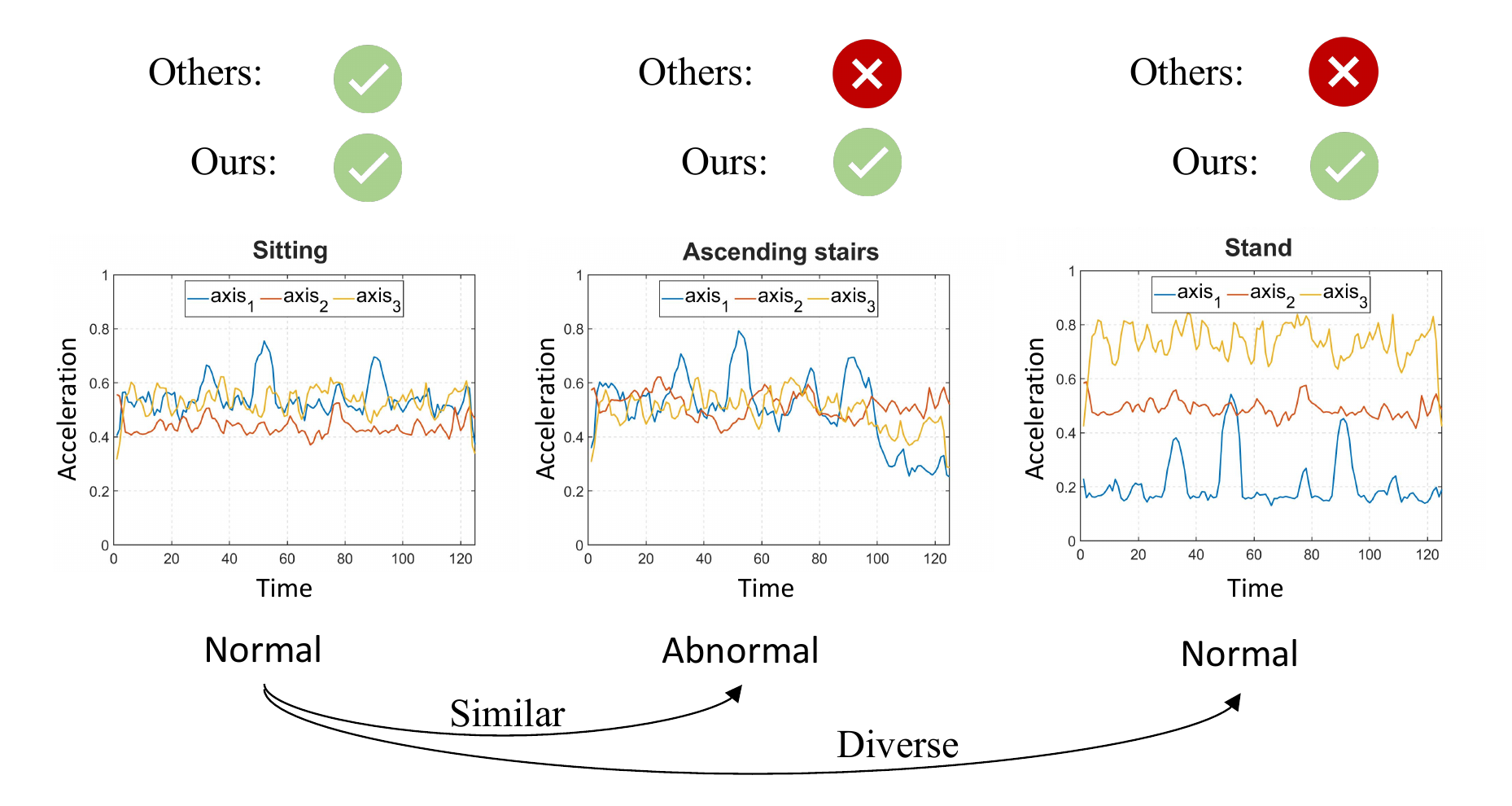}
\caption{Illustration of \method. Even with similar inputs ((a) and (b)), existing methods can overfit to the anomalies; On the other hand, when the inputs are diverse ((a) and (c)), our method remains accurate while existing methods can fail. $Axis$ represents three dimensional signals on dataset.}
\label{fig-motiv}
\end{figure}

Despite the progress, the generalization ability of \uad methods is still limited due to two critical challenges.
The first challenge is limited normal data. 
The normal training data are rather limited compared to the unseen test data, making the model prone to \emph{overfitting}.
As shown in \figurename~\ref{fig-motiv}, when the normal (\figurename~\ref{fig-motiv}(a)) and abnormal samples (\figurename~\ref{fig-motiv}(b)) are similar, existing methods can overfit to the anomalies.
The second challenge is limited feature representations.
For complex time series data, the feature representations learned by existing models lack representativeness required to preserve the diversity of normal patterns.
This is shown in \figurename~\ref{fig-motiv}, when the testing data is becoming diverse to the normal samples (\figurename~\ref{fig-motiv}(c)), existing methods fail to capture the diverse patterns.
Therefore, since the normal training data and feature representations are both limited, the generalization performance of the existing models is poor when applied to unseen normal and abnormal data.

In this paper, we propose a novel \methodfull (\method) to increase the generalization ability of \uad by tackling the above two challenges.
First, to cope with the limited normal training data, \method incorporates a self-supervised learning module to learn general patterns from the normal data.
Second, to cope with the limited feature representations, \method introduces an adaptive memory fusion network to learn common and specific features via the global and local memory modules, respectively.
Then, \method adopts an adaptive fusion module to fuse the global and local representations into the final representation, which is used for reconstruction.
Based on the convolutional autoencoder framework, \method can be easily trained in an end-to-end manner.
As shown in \figurename~\ref{fig-motiv}, while other methods fail in face of the two challenges, our \method can correctly perform anomaly detection.

This work makes the following three contributions:
\begin{enumerate}
    \item We propose \method to tackle the limited normal data and feature representation challenges by adopting self-supervised learning and memory networks, respectively.
    \item We propose to learn both the global and local memories to enhance the representation ability and further propose an adaptive memory fusion module to fuse the global and local memories for final representations.
    \item Extensive experiments on four public datasets demonstrate the effectiveness of \method. Specifically on the largest CAP dataset~\cite{terzano2002atlas} with 900M+ instances, \method significantly outperforms the best comparison approach by \textbf{4}\%+ in both accuracy and F1 score. Moreover, \method is also robust against noise and remain both time and memory efficient.
\end{enumerate}


\section{Related Work}
\label{sec:related}
Unsupervised anomaly detection has been studied for decades. Traditional approaches include reconstruction methods (e.g., PCA, Kernel PCA \cite{paffenroth2013space,Hoffmann2007Kernel}), clustering methods (e.g., GMM, K-means \cite{laxhammar2009anomaly,latecki2007outlier}), and one-class learning methods (e.g., OCSVM, SVDD \cite{banerjee2006support,ma2003time}). Deep learning methods are also popular. They can be categorized into the reconstruction-based and the prediction-based methods. 

\textbf{Reconstruction-based methods} focus on reducing the expected reconstruction errors. For instance, Autoencoders (AEs) \cite{sakurada2014anomaly} are often utilized for anomaly detection by learning to reconstruct a given input. As the model is trained exclusively on normal data, whenever it is not able to reconstruct a given input with equal quality compared to the reconstruction of normal data, the instance is treated as an anomaly. The LSTM Encoder-Decoder model \cite{malhotra2016lstm} is proposed to learn temporal representations of the time series via the LSTM networks and use reconstruction errors to detect anomalies. Despite its effectiveness, LSTM does not take spatial correlation into consideration. Convolutional Autoencoder (CAE) \cite{gutoski2017detection,hasan2016learning} is an important method for video anomaly detection. It is able to capture the 2D image structure since its weights are shared among all locations in the input image. Furthermore, since the Convolutional LSTM (ConvLSTM) can model spatio-temporal correlations by using the convolutional layers instead of the fully connected layers, later works \cite{zhang2019deep,medel2016anomaly} add the ConvLSTM layers into the autoencoder, which encodes the change of appearance for normal data more effectiveness. Others like Variational Autoenocders (VAEs) \cite{an2015variational,xu2018unsupervised}, De-noising AutoEncoders (DAEs) \cite{xu2017detecting}, Deep belief networks (DBNs) \cite{wulsin2010semi}, Replicating neural network \cite{hawkins2002outlier} and Robust Deep Autoencoder (RDA) \cite{zhou2017anomaly} have also reported promising performance. 

\textbf{Prediction-based methods} aim to predict one or more continuous values. In anomaly detection, researchers \cite{lu2017unsupervised,filonov2016multivariate,ergen2017unsupervised} propose RNN-based forecasting models to predict values of the next time period, and minimize the mean squared error between predicted and future values as the criterion for identifying anomalies. There have also been attempts to propose a forecasting model, which uses CNN and RNN, namely LSTNet \cite{lai2018modeling}, to extract short-term local dependency patterns and long-term patterns for multivariate time series analysis for anomaly detection. The GAN-based method \cite{liu2018future} adopts U-Net as the generator to predict the next frame in a video, and leverages adversarial training to distinguish if the predicted result fake (i.e. abnormal events) compared to the ground truth. However, these methods lack a reliable mechanism to learn representations of normal data at a fine-grained level.

Feature representation learning is one important aspect of deep learning and machine learning, and good representations of input data are essential for the generalization ability, interpretability and robustness of methods. Some recent works \cite{li2018deep,zhu2020online} use image context and spatial-temporal relationships to well capture feature correlations for video object detection. \textbf{Self-supervised learning~(SSL)} \cite{jing2020self} is a type of unsupervised learning paradigm that uses the data itself to generate supervision to learn good representations.
For instance, in an image classification problem where most of the images are unlabeled, we can still rotate it by different angles (e.g., $0^{\circ}, 90^\circ, 180^\circ, 270^\circ$), and then use these angles as their labels. In this way, all samples are with labels. By learning on this multi-class classification problem (the auxiliary task), the model can learn general features from these images that can be used for classification (the main task) later. 
There are self-supervised techniques across computer vision \cite{yang2021partially,lin2021completer}, natural language processing \cite{lewis2019bart}, and speech recognition tasks \cite{ravanelli2020multi}. In anomaly detection,  \cite{ali2020self,wang2019effective} use self-supervised visual representation learning  to learn the features of in-distribution (normal) samples.

\textbf{The vanilla memory network} is for question answering~\cite{sukhbaatar2015end}. Generally, the RNNs or LSTM-based models capture long-term structure within sequences through local memory cells. Since the memory in these models is unstable over time, memory networks are proposed by using a global memory with shared read and write functions. Considering that memory can record information stably, several works adopt memory networks such as one-shot learning \cite{cai2018memory,santoro2016meta}, neural machine translation \cite{wang2016memory}, anomaly detection \cite{zhang2020memory,gong2019memorizing,park2020learning}. In anomaly detection, the memory module aims to record various patterns of normal data compared to items in the memory, thus distinguishing between normal and abnormal data.

\section{Proposed Approach}
\label{sec:method}
\begin{figure*}[t!]
\centering
\includegraphics[width = 0.9\textwidth]{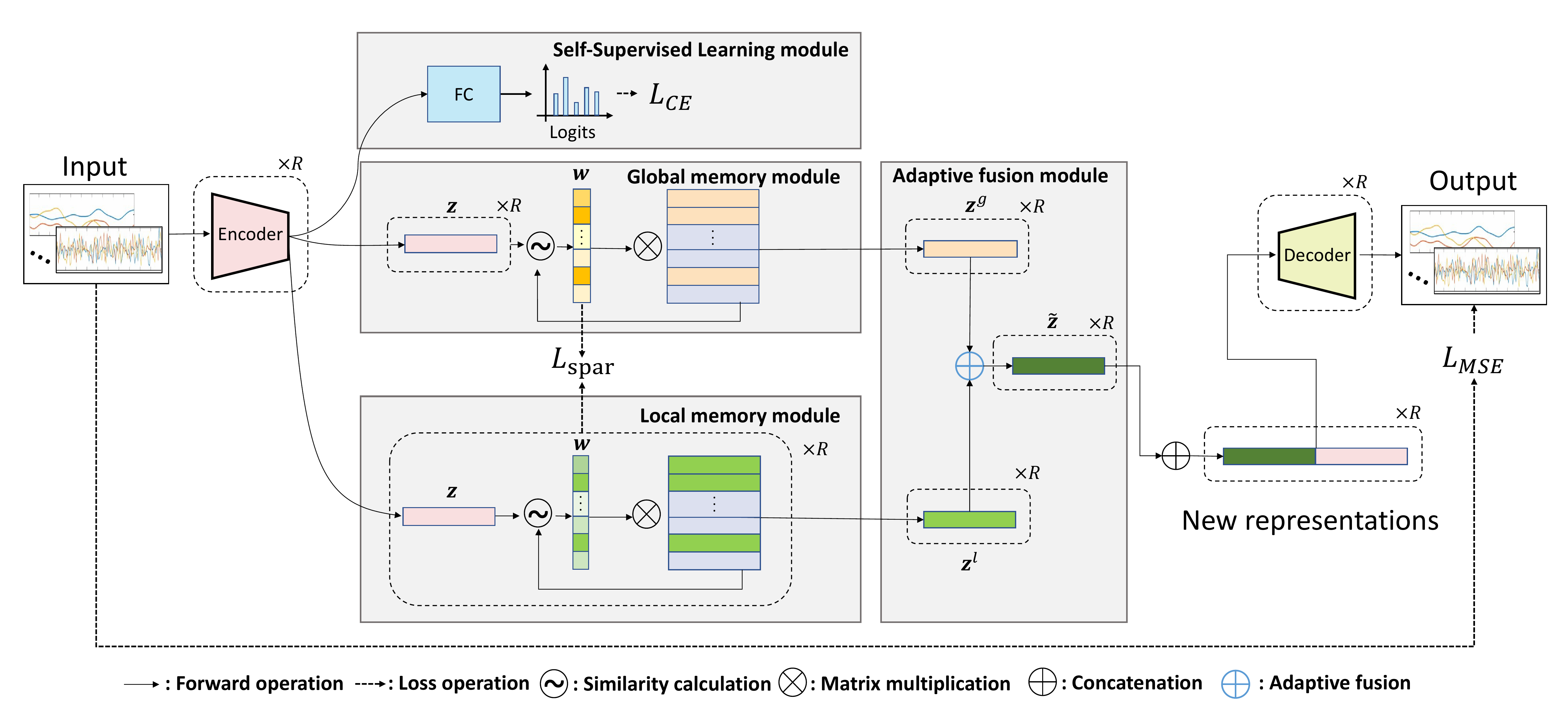}
\caption{The structure of the proposed \method. It consists of four components: self-supervised learning, global memory, local memory and adaptive fusion. The notation ``$\times R$'' denotes $R$ copies where each one corresponds to one transformation.}
\label{fig:overview}
\end{figure*}

\subsection{Problem statement}
\label{sec-method-def}

\begin{definition}[Multivariate time series]
A multivariate time series can be represented as $\bm{X} = (\bm{x}_1, \bm{x}_2, \cdots , \bm{x}_N)^\top \in \mathbb{R}^{N \times V}$, where $\bm{x}_i \in \mathbb{R}^{V}$ is a signal with length $V$ and $N$ is the total number of signals.\footnote{Each sensor signal may have different lengths due to different protocols. In this case, they can still be processed into the same length $V$ by down-sampling or up-sampling.}\footnote{Remark on the data dimension: In real applications, we often employ a sliding window technique to construct samples in time series, thus, $\bm{x}$ is a tensor with $\textgreater 2$ dimensions. For simplicity, we denote its flattened dimension as $V$.}
$y \in \mathcal{Y}$ denotes the corresponding labels, where $\mathcal{Y} = \{1, \cdots, K\}$ represents normal data with $K$ classes.
\end{definition}

\begin{definition}[Anomaly]
A sample $(\bm{x}_a, y_a)$ is called an anomaly if its label $y_a$ does not belong to any predefined classes, i.e. $y_a \notin \mathcal{Y}$.\footnote{Theoretically speaking, there is currently no unified definition of an anomaly.
In this paper, we follow the common definitions in existing literature \cite{shen2020timeseries,zhang2019deep,gong2019memorizing}.}
\end{definition}

In this paper, we deal with the most challenging unsupervised anomaly detection problem, where only unlabeled normal samples are available during training.
This is a more realistic setting since large-scale collection of anomalous samples is often not practical.

\subsection{Overview}
\label{sec-method-overview}
We employ a convolutional autoencoder (CAE) as the base network, which is widely used in existing anomaly detection literature \cite{sakurada2014anomaly,gutoski2017detection,hasan2016learning}.
An autoencoder (AE) is an unsupervised neural network combining an encoder $f_{e}$ and a decoder $f_{d}$ parameterized by $\theta_e$ and $\theta_d$, respectively.
The encoder maps the high-dimensional input $\bm{x} \in \mathbb{R}^V$ into a latent representation $\bm{z} \in \mathbb{R}^{F}$, where $F \ll V$. Then, the decoder $f_{d}$ reconstructs the original input as $\bm{x}^\prime$ by mapping $\bm{z}$ back into the input space.
The latent representation $\bm{z}$ and the reconstructed input $\bm{x}^ \prime$ can be respectively computed as:
\begin{equation}
\begin{split}
    \bm{z} = f_e (\bm{x}; \theta_e),
    \bm{x}^ \prime = f_d(\bm{z};\theta_d).
\end{split}
\end{equation}

The reconstruction error is calculated using the Mean Squared Error (MSE):
\begin{equation}
\label{eq-mse}
L_{MSE} = \lVert \bm{x} - \bm{x}^ \prime \lVert ^2_2,
\end{equation}
where $\lVert \cdot \lVert^2_2$ is the $\textit{l}_2$-norm.


In this paper, we propose a novel \methodfull (\method) for \uad.
\method consists of four novel components as shown in \figurename~\ref{fig:overview}: 1) a self-supervised learning module, 2) a global memory module, 3) a local memory module and 4) an adaptive fusion module. \method works in the following four steps.
\begin{enumerate}
    \item First, the encoder maps the raw time series signal and its six transformations into a latent feature space. 
    \item Then, for self-supervised learning, a multi-class classifier is built to classify these feature types in order to learn generalized feature representations.
    \item Meanwhile, the features are also fed into the global and local memory network modules to learn both common and specific features. 
    \item Finally, the adaptive fusion module fuses these features to obtain new representation which is used for reconstruction.
\end{enumerate}
Details of \method are presented in the following sections.

\subsection{Self-Supervised learning}
In this section, we introduce the self-supervised learning module of \method that enables generalized feature representation learning for normal data.
Compared to the potentially large number of unseen anomalies that could be in any forms of representations, the number of the normal training data is relatively limited. Hence, anomaly detection models trained on such limited normal samples tend to overfit.
However, it is prohibitively expensive to collect all the infinite training data.
To address this issue, we propose to use self-supervised learning to increase the model's generalization ability.

Assuming that the instances are consistent before and after basic data augmentations (i.e., normal data and abnormal data are still distinguishable after the same augmentations)~\cite{jing2020self}, we design feature transformations on the original data for self-supervision.
Then, we train the model to recognize the transformation type of a sample as its auxiliary task. In the following experiments, as shown in \figurename~\ref{fig-sub-ssl}, we can observe that the instances after the same augmentations can still identify normal and abnormal data.
Specifically, we propose to utilize six signal transformations inspired by~\cite{saeed2019multi}, which are described as follows: 
\begin{enumerate}
  \item \emph{Noise:} As noisy sensor signal may exist in the real world, adding noise to signal can help model learn more robust features against the noise. Here, the transformation with Gaussian noise is implemented.
  \item \emph{Reverse:} This transformation reverses the samples along the temporal dimension, resulting in the samples with the opposite time direction.
  \item \emph{Permute:} This transformation randomly perturbs signal along the time dimension by slicing and swapping different time windows to generate new samples. It aims to enhance the permutation invariant property of the resulting model.
  \item \emph{Scale:} Scaling changes the magnitude of signals within a time window by multiplying a random scalar. Here, we select $[0.5,0.8,1.5,2]$ as scalar values. The addition of scaled signals can help the model learn scale-invariant patterns.
  \item \emph{Negate:} This transformation is a special type of scaling transformations. It is scaled by -1, resulting in a mirror image of the input signals.
  \item \emph{Smooth:} This transformation applies the Savitzky-Golay (SG) method to smooth the signals. The Savitzky-Golay filter is a particular type of low-pass filters, well adapted for noisy signal smoothing.
\end{enumerate}

To learn general feature representations from these transformations, the model is enforced to distinguish their transformation types by using the Cross-Entropy loss function:
\begin{equation}
\label{eq-ssl-cross-entropy}
L_{CE} = -\sum_{i=1}^{R} y_{i} \log \left(p_{i}\right),
\end{equation}
where $R$ denotes the number of self-supervised learning classes ($R=7$ in this work by including all six transformations and the original signal).
$y_i$ and $p_i$ are the pseudo label and the predicted probability, respectively.
A Softmax activation function is applied to the probability before the cross-entropy loss.


\subsection{Adaptive Memory Fusion Module}
\label{sec-method-memory}
Traditional AE is negatively affected by noisy or unknown training data, so it can also consistently reconstruct abnormal inputs too well \cite{zong2018deep,gong2019memorizing}.
Therefore, the model fails to learn representative features.
To tackle this challenge, we propose a \emph{adaptive memory fusion module} to enhance the ability of the model in distinguishing between normal and abnormal data via recording the prototypical patterns.
In the following, we will first introduce the memory network.
Then, we present our adaptive memory fusion module in detail.

\subsubsection{Memory module}
The memory module \cite{zhang2020memory,gong2019memorizing} consists of a memory representation to represent the encoded patterns, and a memory updating part to update the memory items based on the similarity of the memory items and the input $\bm{z}$.
Specifically, the memory is instantiated as a matrix $\bm{M} \in \mathbb{R}^{ C \times F}$, storing $C$ vectors of dimension $F$. Let the row vector $\bm{m}_i$, $\forall i \in [C]$ denote the $i$-th row of $\bm{M}$, where $[C]$ denotes the set of integers from $1$ to $N$. Each $\bm{m}_i$ denotes a memory item.
Given a query $\bm{z} \in \mathbb{R} ^{F}$ (i.e. encoding), where $F$ denotes the filter size in the last layer of the encoder,
the memory-guided module outputs $\bm{\hat{z}}$, which represents a weighted average of memory items $\bm{m}_i$, by matching probabilities $\bm{w} \in \mathbb{R}^{C}$ between the query $\bm{z}$ and memory matrix $\bm{M}$ as follows:
\begin{equation}
\label{z6}
\begin{aligned}
\hat{\bm{z}} = \bm{w M} = \sum^C_{i=1} w_i \bm{m}_i,
\end{aligned}
\end{equation}
where $w_i$ denotes the $i$-th entry of $\bm{w}$. The weight vector $\bm{w}$ is computed based on the normalized similarity between the query $\bm{z}$ and memory entry $\bm{m}_i$:
\begin{equation}
w_i=\frac{\exp \left(\operatorname{Score}\left(\bm{z}, \bm{m}_i \right)\right)}{\sum_{j=1}^{C} \exp \left(\operatorname{Score}\left(\bm{z}, \bm{m}_j \right)\right)}.
\end{equation}

The $\operatorname{Score}$ function is implemented as the cosine similarity:
\begin{equation}
\operatorname{Score}\left(\bm{z}, \bm{m}_i \right)=\frac{\bm{z} \cdot \bm{m}_i^{\top}}{\lVert \bm{z} \lVert \lVert \bm{m}_i \lVert}.
\end{equation}

In the training phase, the memory matrix can be updated by the reconstruction loss function, thus forcing it to record the normal characteristics.
In the testing phase, the memory network outputs a representation with a combination of all items, taking into account the multiple patterns of normal characteristics.
Therefore, the normal instances can be reconstructed well.
The anomalies reconstructed using the retrieved normal patterns in the memory module will result in higher reconstruction errors.

\subsubsection{Adaptive fusion module}

We further propose an adaptive memory fusion network to learn both the common and specific representations from all the feature augmentations. 
Specifically, we propose the \emph{global} memory module to learn the common representations contained in all transformations, and the \emph{local} memory module to learn augmentation-specific representations for each transformation.
Finally, we propose a \emph{adaptive} fusion module to fuse these two levels of features into the final representations that will be used for reconstruction.
The motivation is that we can capture the common patterns for the normal data and its specific information (i.e. each different transformation) that is useful for the normal data patterns, thus improving feature representations of normal data at a fine-grained level.

We build the global memory module with a shared memory matrix.
By using the encoded representation as a query, the global memory module can record general items in the memory matrix.
Through a shared memory module, the outputs are obtained as:
\begin{equation}
\label{eq-mem-global}
\begin{aligned}
\bm{z}^g_i  = f_{g}(f_e(\bm{x}_i;\theta _e);\theta_g), i \in [ R ],
\end{aligned}
\end{equation}
where $f_{g}(\cdot )$ is the function of the global memory module. $[ R ] = \{1, 2, \cdots, R\}$ and $\theta_g$ denote the shared parameters of one global memory module.  

We build $R$ local memory modules for the raw data and the six transformations.
Each memory matrix records the corresponding normal characteristics of transformation.
These outputs are obtained by the local memory modules as:
\begin{equation}
\label{eq-mem-local}
\begin{aligned}
\bm{z}^l_i = f_{l}(f_e(\bm{x}_i;\theta _e);\theta_l^i), i \in [ R ],
\end{aligned}
\end{equation}
where $f_l(\cdot)$ is the function of the local memory module, and $\theta_l^i$ denotes the parameters of seven local memory modules.

It is intuitive that the common and specific features are not equally important in representing a given instance.
To adaptively fuse these features, we use a feed-forward layer that takes the features and a free variable $r \in \mathbb{R}$ as inputs to generate the fused representations by weights $\bm{\alpha} \in \mathbb{R}^{2R}$ ($2$ weights for local and global memories, with $R$ transformations in total).
Note that we use Batch Normalization and sigmoid activation to normalize the weights and control their values to be within the range of $(0,1)$.
$r$ is used to increase randomness.
Then, we get the adaptive fused representations:
\begin{equation}
\label{eq-fusion}
\begin{aligned}
\tilde{\bm{z}_i}= \begin{bmatrix} \alpha^g_i&\alpha^l_i \end{bmatrix} \begin{bmatrix}
\bm{z}^g_i  \\ \\ \bm{z}^l_i
\end{bmatrix}, i \in [ R ],
\end{aligned}
\end{equation}
where $\alpha^g_i, \alpha^l_i \in \mathbb{R}$ denote the weights of common (global) and specific (local) features, respectively.

The decoders concatenate $\bm{z}$ (encoding output) and $\tilde{\bm{z}}$ (adaptive fusion output) as inputs to reconstruct the original inputs. The reconstruction loss is defined by minimizing the $l_{2}$ distance between the decoder output and the original input as follows:
\begin{equation}
\label{eq-mse}
\begin{aligned}
L_{MSE} = \sum^R_{i=1} \lVert f_d(\operatorname{concat}(\tilde{\bm{z}_i},\bm{z}_i);\theta_d^i)- \bm{x}_i \lVert ^2_2.
\end{aligned}
\end{equation}

To constrain the sparsity of the memory weight $\bm{w}$ to avoid over-reconstruction of anomalies by the complex combination of memory items, we adopt sparse loss by minimizing the entropy of $\bm{w}$:
\begin{equation}
\label{eq-sparse}
\begin{aligned}
L_{spar} = \sum_{i=1}^{C}-w_i \cdot log(w_i).
\end{aligned}
\end{equation}

\subsection{Training and Inference}
\subsubsection{Training}

By integrating the reconstruction loss in Eq.~\eqref{eq-mse}, the sparse loss in Eq.~\eqref{eq-sparse}, and the self-supervised loss in Eq.~\eqref{eq-ssl-cross-entropy} with tradeoff parameters $\lambda_1, \lambda_2$, we obtain the overall training objective for \method which can be optimized in an end-to-end manner:
\begin{equation}
\label{z19}
\begin{split}
J(\theta) = L_{MSE} + \lambda_1 L_{CE}+ \lambda_2 L_{spar} . 
\end{split}
\end{equation}

The training procedure of \method on multivariate time series is shown in Algorithm \ref{algorithm1}. Note that we need to take of the multiple dimensions of the inputs in real implementation.

\subsubsection{Inference}
\label{threshold}
\renewcommand{\algorithmicrequire}{\textbf{Input:}}  
\renewcommand{\algorithmicensure}{\textbf{Output:}}  
\begin{algorithm}[t!]
	\caption{Learning procedure of \method}
	\begin{algorithmic}[1]
		\REQUIRE  Normal Dataset $\bm{\mathcal{X}} = \{\bm{X}_1, \bm{X}_2, \cdots, \bm{X}_H \}$, hyperparameters $\lambda_1$, $\lambda_2$, variable $r$.
		\ENSURE Decision threshold $\mu$ and model parameter $\theta$.
        \STATE Transform each sample into  $\bm{X} = \{\bm{x}^1,...,\bm{x}^R\} \in \mathbb{R}^{R \times V \times N}$, where $R = 7$ including raw data and 6 transformations;
        \STATE Initialize all parameters;
		\WHILE{not converge} 
		\STATE $\bm{Z} \leftarrow Encoder(\bm{X})$; 
		\STATE $p \leftarrow FC(\bm{Z})$; 
		\STATE $\bm{Z}^l,\bm{w} \leftarrow LMem(\bm{Z})$; 
		\STATE $\bm{Z}^g,\bm{w} \leftarrow GMem(\bm{Z})$; 
		\STATE Learn adaptive fuse weights $\bm{\alpha}$;
		\STATE $\tilde{\bm{Z}} \leftarrow Fusion(\bm{Z}^l,\bm{Z}^g;\bm{\alpha})$; 
		\STATE $\bm{X}^\prime \leftarrow Decoder( Concat(\tilde{\bm{Z}},\bm{Z}))$;
		\STATE $J(\theta) = \frac {1} {H} \sum_{i=1}^{H}  ( \sum^R_{j=1} L(\bm{X}^\prime_{ij},\bm{X}_{ij})- \lambda_1 \sum_{j=1}^{R} y_{ij} \log \left(p_{ij}\right) -\lambda_2 \sum_{j=1}^{2R} \sum_{k=1}^{C}w_{ijk} \cdot log(w_{ijk}))$; //Implementation of Eq.~\eqref{z19} for multivariate time series
		\STATE Update model parameter $\theta$ using Eq.~\eqref{z19};
		\ENDWHILE
		\STATE Calculate the decision threshold $\mu$ by the training samples;
		\RETURN Optimal $\mu$ and $\theta$.
	\end{algorithmic} 
	\label{algorithm1}
\end{algorithm}

For the autoencoder-based models, it is generally assumed that the compression is different for instances of different categories.
That is, if the training dataset only contains normal instances, the reconstruction error becomes higher for abnormal instances. Therefore, we can classify these instances into ``abnormal'' or ``normal'' by their reconstruction error during inference phase.


Given normal samples as the training dataset $\bm{\mathcal{X}} = \{\bm{X}_1, \cdots, \bm{X}_H\}$. First, we need to construct the self-supervised transformations. 
Let the corresponding decision threshold $\mu$ be the 99th percentile of $\mathrm{Err}(\bm{X}_i)$ on this training set, where $\mathrm{Err}(\bm{X}_i)$ is the value of reconstruction loss function $L_{MSE}$ for $\bm{X}_i$.
In inference (or detection) process, the decision rule is that if $\mathrm{Err}(\bm{X}_i)>\mu $, the testing sample $\bm{X}_i$ in a sequence will be classified as ``abnormal''; otherwise, it will be classified as ``normal''. Here, LMem and GMem denote local memory module and global memory module, respectively. The inference procedure of \method is shown in Algorithm~\ref{algorithm2}.

\renewcommand{\algorithmicrequire}{\textbf{Input:}}  
\renewcommand{\algorithmicensure}{\textbf{Output:}}  
\begin{algorithm}[t!]
	\caption{Inference procedure of AMSL}
	\begin{algorithmic}[1]
		\REQUIRE  Testing dataset $\bm{\mathcal{X}} = \{\bm{X}_1, \bm{X}_2, \cdots, \bm{X}_H\}$, $\bm{X} \in \mathbb{R}^{V \times N}$, threshold $\mu$, model parameter $\theta $, hyperparameters $\lambda_1$, $\lambda_2$.
		\ENSURE Label $y_i$ of $\bm{X}_i$.
		\STATE Transform each sample into  $\bm{X} = \{\bm{x}^1,...,\bm{x}^R\} \in \mathbb{R}^{R \times V \times N}$, where $R = 7$ including raw data and 6 transformations;
        \STATE Load the model trained parameters $\theta $;
		\STATE $\bm{Z} \leftarrow Encoder(\bm{X})$; 
		\STATE $\bm{Z}^l \leftarrow LMem(\bm{Z})$; 
		\STATE $\bm{Z}^g \leftarrow GMem(\bm{Z})$; 
		\STATE Load $\bm{\alpha}$;
		\STATE $\tilde{\bm{Z}} \leftarrow Fusion(\bm{Z}^l,\bm{Z}^g;\bm{\alpha})$; 
		\STATE $\bm{X}^\prime \leftarrow Decoder(Concat(\tilde{\bm{Z}},\bm{Z}))$;
		\STATE Calculate reconstruction loss $Err(\bm{X}_i)= \sum^R_{j=1} L(\bm{X}^\prime_{ij},\bm{X}_{ij})$;
		\IF{$Err(\bm{X}_i)>\mu$} 
		\STATE $y_i$ = ``Abnormal'';
		\ELSE
		\STATE ${y_i}$ = ``Normal'';
		\ENDIF
		\RETURN $y_i$.
	\end{algorithmic} 
	\label{algorithm2}
\end{algorithm}

\section{Experimental Evaluation}
\label{sec:exp}

\subsection{Datasets}

In the experiments, we adopt the following four datasets for evaluation as shown in \tablename~\ref{tb-dataset}.

\begin{table}[htbp]
	\centering
	\caption{The detailed statistics of the four datasets}
	\label{tb-dataset}
	\resizebox{.5\textwidth}{!}{
	\begin{tabular}{llrrr}
		\toprule
		\textbf{Dataset} & \textbf{Application} & \textbf{\#Instance} & \textbf{\#Dim} & \textbf{\#Class} \\ \hline 
		DSADS \cite{altun2010comparative} & Activity Recognition & 1,140,000 & 45 & 19  \\
		PAMAP2 \cite{reiss2012introducing} & Activity Recognition & 2,844,868 & 36 & 18\\ 
		WESAD \cite{schmidt2018introducing} & Stress \& Affect Detection & 63,000,000 & 14 & 3 \\ 
		CAP \cite{terzano2002atlas} & Sleep Stage Detection & 921,700,000 & 21 & 8 \\ 
         \bottomrule
	\end{tabular}
	}
\end{table}

\emph{DSADS}~\cite{altun2010comparative} consists of motion sensor data (i.e. accelerometer, gyroscope, magnetometer) of 19 activities of daily living and sports activities performed by 8 subjects.
To simulate normal and abnormal classes, we chose running, ascending stairs, descending stairs, rope jumping and playing basketball as anomaly classes, while the remaining categories are regarded as normal classes.
These designated anomaly activities are relatively intense and rare compared to other activities. In \tablename~\ref{table-class1}, we describe the selection of normal and abnormal classes on DSADS datasets. We chose these activities as the abnormal classes that are relatively intense and rare activities compared to other activities, while the rest categories are defined as the normal classes.

\emph{PAMAP2}~\cite{reiss2012introducing} is a mobile dataset containing data of 18 different physical activities performed by 9 subjects wearing 3 inertial measurement units comprising an accelerator, a gyroscope and a magnetometer.
As shown in \tablename~\ref{table-class2}, we treat the classes with relatively small samples as the anomaly classes, including running, ascending stairs, descending stairs and rope jumping. The remaining categories are used as the normal classes in experiments.

\begin{table}[t!]
    \caption{The selection of normal and abnormal classes on DSADS and PAMAP2 datasets}
    \centering
\subtable[DSADS]{
\begin{tabular}{cc}
\toprule
Activity                                                                                                                  & Class    \\ \hline
\begin{tabular}[c]{@{}c@{}}sitting, standing, lying, moving, \\ walking, cycling, exercising, rowing\end{tabular}         & normal   \\ \hline
\begin{tabular}[c]{@{}c@{}}running, ascending stairs, descending stairs, \\ rope jumping, playing basketball\end{tabular} & abnormal \\  \bottomrule
\end{tabular}
\label{table-class1}
}
 
\qquad
 
\subtable[PAMAP2]{        
\begin{tabular}{cc}
\toprule

Activity                                                                                                                        & Class    \\ \hline
\begin{tabular}[c]{@{}c@{}}lying, sitting, standing, walking, cycling, \\ Nordic walking, vacuum cleaning, ironing\end{tabular} & normal   \\ \hline
\begin{tabular}[c]{@{}c@{}}running, ascending stairs, descending stairs, \\ rope jumping\end{tabular}                           & abnormal \\  \bottomrule
\end{tabular}
\label{table-class2}
}
\end{table}

\begin{table*}[t!]
	
	\centering  
	\caption{The comparison of mean precision, recall, F1 and accuracy of \method and other baselines. The best and second-best results are bold and underlined, respectively. We can see that \method significantly outperforms other methods.}  
	\label{tab:performance_comparison}
	\resizebox{1\textwidth}{!}{
	\begin{tabular}{ccccccccccccccccc}  
			\toprule  
			\multirow{2}{*}{Method}&\multicolumn{4}{c}{DSADS dataset}& 
			\multicolumn{4}{c}{PAMAP2 dataset}&\multicolumn{4}{c}{WESAD dataset}&\multicolumn{4}{c}{CAP dataset}\cr  
			\cmidrule(lr){2-5} \cmidrule(lr){6-9}\cmidrule(lr){10-13}\cmidrule(lr){14-17}      
			&mPre&mRec&mF1&Acc&mPre&mRec&mF1&Acc&mPre&mRec&mF1&Acc&mPre&mRec&mF1&Acc\cr  
			\midrule  
			Kernel PCA~\cite{Hoffmann2007Kernel}&0.6184&0.6182&0.6183&0.6186&0.7236&0.6579&0.6892&0.5645&0.5496&0.5495&0.5495&0.5486&0.7603&0.5847&0.6611&0.5892\cr  
			ABOD~\cite{kriegel2008angle}&0.6880&0.6510&0.6690&0.6554&0.8653&0.9022&0.8834&0.8985&0.8782&0.8786&0.8784&0.8783&0.7867&0.6365&0.7037&0.6326\cr  
			OCSVM~\cite{ma2003time}&0.7608&0.7277&0.7439&0.7312&0.7600&0.7204&0.7397&0.7679&0.6092&0.5631&0.5852&0.5518&\underline{0.9267}&\underline{0.9259}&\underline{0.9263}&\underline{0.9257}\cr  
			HMM~\cite{joshi2005investigating}&0.6959&0.6917&0.6937&0.6901&0.6950&0.6553&0.6745&0.5725&0.6123&0.6060&0.6097&0.6018&0.8238&0.8078&0.8157&0.8090\cr  
			\hline  
			  
			CNN-LSTM~\cite{donahue2015long}&0.6845&0.6270&0.6545&0.6425&0.6680&0.5392&0.5968&0.6131&0.5883&0.5440&0.5653&0.5318&0.6159&0.5217&0.5649&0.5762\cr  
			LSTM-AE~\cite{malhotra2016lstm}&0.8471&0.7729&0.8083&0.7852&0.8619&0.7997&0.8296&0.8426&0.2353&0.4762&0.3150&0.4599&0.7147&0.6253&0.6671&0.6286\cr 
			
			MSCRED~\cite{zhang2019deep}&0.7540&0.5602&0.6428&0.6192&0.6997&0.7301&0.7146&0.7517&0.8850&0.8124&0.8471&0.8420&0.6410&0.5784&0.6081&0.5819\cr 
			ConvLSTM-AE~\cite{medel2016anomaly}&0.8164&0.6951&0.7509&0.7121&0.7359&0.7361&0.7360&0.7341&\underline{0.9733}&\underline{0.9698}&\underline{0.9716}&\underline{0.9709}&0.8150&0.8194&0.8172&0.8165\cr  
			ConvLSTM-COMP~\cite{medel2016anomaly}&0.8229&0.7379&0.7781&0.7518&\underline{0.8844}&\underline{0.8842}&\underline{0.8843}&\underline{0.8844}&0.9626&0.9629&0.9627&0.9619&0.8367&0.8377&0.8372&0.8394\cr 
			BeatGAN~\cite{zhou2019beatgan}&0.9517&0.5663&0.7100&0.7818&0.7981&0.7420&0.7691&0.8369&0.7586&0.5000&0.6027&0.5172&0.5251&0.5002&0.5123&0.8437\cr
			MNAD-P~\cite{park2020learning} &0.5816&0.5783&0.5799&0.5721&0.8198&0.8176&0.8186&0.8135&0.7600&0.6938&0.7254&0.6849&0.7742&0.7489&0.7613&0.6960\cr 
			MNAD-R~\cite{park2020learning} &0.8337&0.7694&0.8003&0.7811&0.8350&0.8355&0.8353&0.8334&0.7426&0.6677&0.7031&0.6579&0.8189&0.8235&0.8212&0.7871\cr 
			
			GDN~\cite{deng2021graph}&0.8706&0.8151&0.8419&0.8251&0.8129&0.8104&0.8116&0.8123&0.7520&0.5434&0.6309&0.5590&0.6831&0.6237&0.6520&0.6569\cr
			
			UODA~\cite{lu2017unsupervised}&\underline{0.8679}&\underline{0.8281}&\underline{0.8475}&\underline{0.8365}&0.8957&0.8513&0.8730&0.8823&0.7623&0.6314&0.6907&0.6191&0.7557&0.5124&0.6107&0.5173\cr
			
			\hline 
			
			\method (Ours)&{\bf 0.9407}&{\bf 0.9298}&{\bf 0.9352}&{\bf 0.9332}&{\bf 0.9788}&{\bf 0.9713}&{\bf 0.9750}&{\bf 0.9770}&{\bf 0.9953}&{\bf 0.9949}&{\bf 0.9951}&{\bf 0.9951}&{\bf 0.9771}&{\bf 0.9736}&{\bf 0.9753}&{\bf 0.9756}\cr
			 
			Improvement &{ 7.28\%}&{ 10.17\%}&{ 8.77\%}&{ 9.67\%}&{ 9.44\%}&{ 8.71\%}&{ 9.07\%}&{ 9.26\%}&{ 2.20\%}&{ 2.51\%}&{ 2.35\%}&{ 2.42\%}&{ 5.04\%}&{ 4.77\%}&{ 4.90\%}&{ 4.99\%}\cr
			\bottomrule  
		\end{tabular}}
\end{table*} 

\textit{WESAD}~\cite{schmidt2018introducing} is a public dataset for wearable stress and affect detection, which contains physiological and motion data of 15 subjects. We use sensor modalities of chest-worn device including electrocardiogram, electrodermal activity, electromyogram, respiration, body temperature, and three-axis acceleration. In our experiments, we chose normal emotional states (neutral, amusement) as normal classes and stress states as anomaly class.

\textit{CAP Sleep Database}~\cite{terzano2002atlas}, which stands for the Cyclic Alternating Pattern (CAP) database. It is a clinical dataset from PhysioNet repository \cite{cap}.
It is characterized by periodic physiological signals occurring during wake, S1-S4 and REM sleep stages.
The signals include EEG, EOG, EMG, EKG and respiration.
There are 16 healthy subjects and 92 patients in the database.
In this task, we extracted 7 valid channels (ROC-LOC, C4-P4, C4-A1, F4-C4, P4-O2, ECG1-ECG2 and EMG1-EMG2).
For detecting sleep apnea events, we chose healthy subjects as the normal class and the patients with sleep-disordered breathing as the anomaly class.

\subsection{Comparison Methods} 
We compare \method against four popular traditional anomaly detection methods, which can be applied after feature extraction:

\begin{itemize}
\item \emph{KPCA}~\cite{Hoffmann2007Kernel}, which is a non-linear extension of PCA commonly used for anomaly detection. We adopt Gaussian kernel in our experiments.
\item \emph{ABOD}~\cite{kriegel2008angle}, which uses $k$ nearest neighbors to approximate the complexity reduction. For an observation, the variance of its weighted cosine scores to all neighbors could be viewed as the abnormal score.
\item \emph{OCSVM}~\cite{ma2003time}, which adopts PCA for dimension reduction and employs the Gaussian kernel with a bandwidth of $0.1$.
\item \emph{HMM}~\cite{joshi2005investigating}, which is applied after extracting features and then it calculates the anomaly probability from the state sequence generated by the model.
\end{itemize}

We also compare it against seven deep unsupervised methods: 
\begin{itemize}
\item \emph{CNN-LSTM}~\cite{donahue2015long}, which is a prediction-based method built by firstly defining a network comprised of Conv2D and MaxPooling2D layers ordered into a stack of the required depth. The result is then fed into the LSTM and FC layers as prediction.
\item \emph{LSTM-AE}~\cite{malhotra2016lstm}, which is a reconstruction-based method using a single-layer LSTM on both encoder and decoder.
\item \emph{MSCRED}~\cite{zhang2019deep}, which is an encoder-decoder model with multi-scale matrices as inputs for multivariate time series analysis.
\item \emph{ConvLSTM-COMPOSITE}~\cite{medel2016anomaly}, which is a composite encoder-decoder model with reconstruction and prediction task. We choose the ``conditional'' version to build a single model called \emph{ConvLSTM-AE} by removing the forecasting decoder.
\item \emph{BeatGAN}~\cite{zhou2019beatgan}, which is a reconstruction-based method with an adversarial generation approach as regularization.
\item \emph{MNAD}~\cite{park2020learning}, which is an encoder-decoder model based on a memory module for video anomaly detection. It has two variants: one with the prediction task (\emph{MNAD-P}), another with the reconstruction task (\emph{MNAD-R}).
\item \emph{GDN}~\cite{deng2021graph}, which is a Graph-based neural network to learn a graph of the dependence relationships between sensors for anomaly detection. 
\item \emph{UODA}~\cite{lu2017unsupervised}, which is a RNN-based network for anomaly detection. We re-implemented it by customizing the number of layers and hyper-parameters.
\end{itemize}

We re-implement the comparison methods based on several open-source repositories\footnote{\url{https://pyod.readthedocs.io/en/stable/}, \url{https://github.com/7fantasysz/MSCRED},\url{https://github.com/cvlab-yonsei/MNAD},\url{https://github.com/Vniex/BeatGAN}, \url{https://github.com/d-ailin/GDN}} as well as our own implementations.
As for ours, there is a data pre-processing stage where data is normalized, split into windows with length $V$ and transformed. The encoder runs with $Conv2D$ $\rightarrow$ $ Maxpool$ $\rightarrow$ $Conv2D$ $\rightarrow$ $Maxpool$, i.e., Conv1- Conv2 with $32$ and $64$ kernels of size $ 4 \times 4$, and Maxpooling with size $2 \times 2$. In adaptive memory fusion module, $F=64$ and $C = 800$ are the best choice.
Besides, we use a variable $r$ as the initial weight of adaptive fusion network with $FC(2R)$ $\rightarrow$ $BN$ $\rightarrow$ $Sigmoid$ $\rightarrow$ $Multiply$. These weights are multiplied into local and global feature representations by Eq. \eqref{eq-fusion}.
As the input of the decoder is by concatenating the output of encoder and the output of memory module, the decoder is asymmetric to the encoder that uses 4 $Conv2DTranspose$ modules, with $\{128, 64, 32, 1\}$ kernels of size $4\times 4$ in each layer, respectively. To compute classification error, the output of encoder is also incorporated into classification network with $Conv2D$ $\rightarrow$ $Flatten$ $\rightarrow$ $FC(128)$ $\rightarrow$ $Dropout$ $\rightarrow$ $FC(R)$, i.e. Conv with 1 kernel of size $4\times 4$. 
\method is trained in an end-to-end fashion using \texttt{Keras} \cite{ketkar2017introduction} on a TITAN XP GPU. The Adam optimizer is used to train model for about 100 epochs with 32 or 64 batch size. The learning rate is 0.001. And we set the hyper-parameters: $\lambda_1 = 1$ and $\lambda_2 = 0.0002$ while parameter sensitivity analysis is presented in later sections.


In practice, it is difficult to know the ground
truth, and anomalous data points are rare. Hence, the semi-supervised setting is a commonly used as evaluation method \cite{gong2019memorizing,zhai2016deep,zong2018deep}, the training set only contains the normal samples and has no overlapping with the testing set.
For each dataset, we split the normal samples into training, validation, and testing sets with the ratio of $5:1:4$. The model selection criterion, i.e., hyperparameters, used for tuning is the validation error on the validation set.
Besides, since most datasets have more normal samples than anomalies, accuracy along is insufficient for evaluation.
Thus, for comprehensive evaluation, we adopt four evaluation metrics: mean \emph{precision, recall}, \emph{F1} score, and \emph{accuracy} following existing literature~\cite{zong2018deep,lu2017unsupervised,zhai2016deep}.

\subsection{Results and Analysis}
\tablename~\ref{tab:performance_comparison} reports the overall performance results on these public datasets.
It can be observed that the proposed \method method achieves significantly superior performance over the baseline methods in all the datasets. 
Specifically, compared with other methods, \method significantly improves the F1 score by $\textbf{9.07}\%$ on PAMAP2 dataset, $\textbf{4.90}\%$ on CAP dataset, $\textbf{8.77}\%$ on DSADS dataset and $\textbf{2.35}\%$ on WESAD dataset. The same pattern goes for precision and recall. 
Especially for the largest CAP dataset with over 900 Millon samples, \method dramatically outperforms the second-best baseline (OCSVM) with an F1 score of $\textbf{4.90}\%$, indicating its effectiveness.

For DSADS, PAMAP2 and CAP datasets (WESAD is relatively easier to train than the other datasets), we find that as the datasets become larger, the improvements tend to decrease.
This means that self-supervision is more useful on small-scale datasets where it is hard to learn generalized representations.
This is in accordance with existing machine learning conclusions.
In addition, even if the samples are relatively fewer in DSADS and PAMAP2 datasets but with more categories, our \method still significantly outperforms other methods by large margins, which indicate its ability of dealing with the diversities in the limited training data.


\begin{figure*}[t!]
    \centering
    \subfigure[SSL transformations]{
    \includegraphics[width=.3\textwidth]{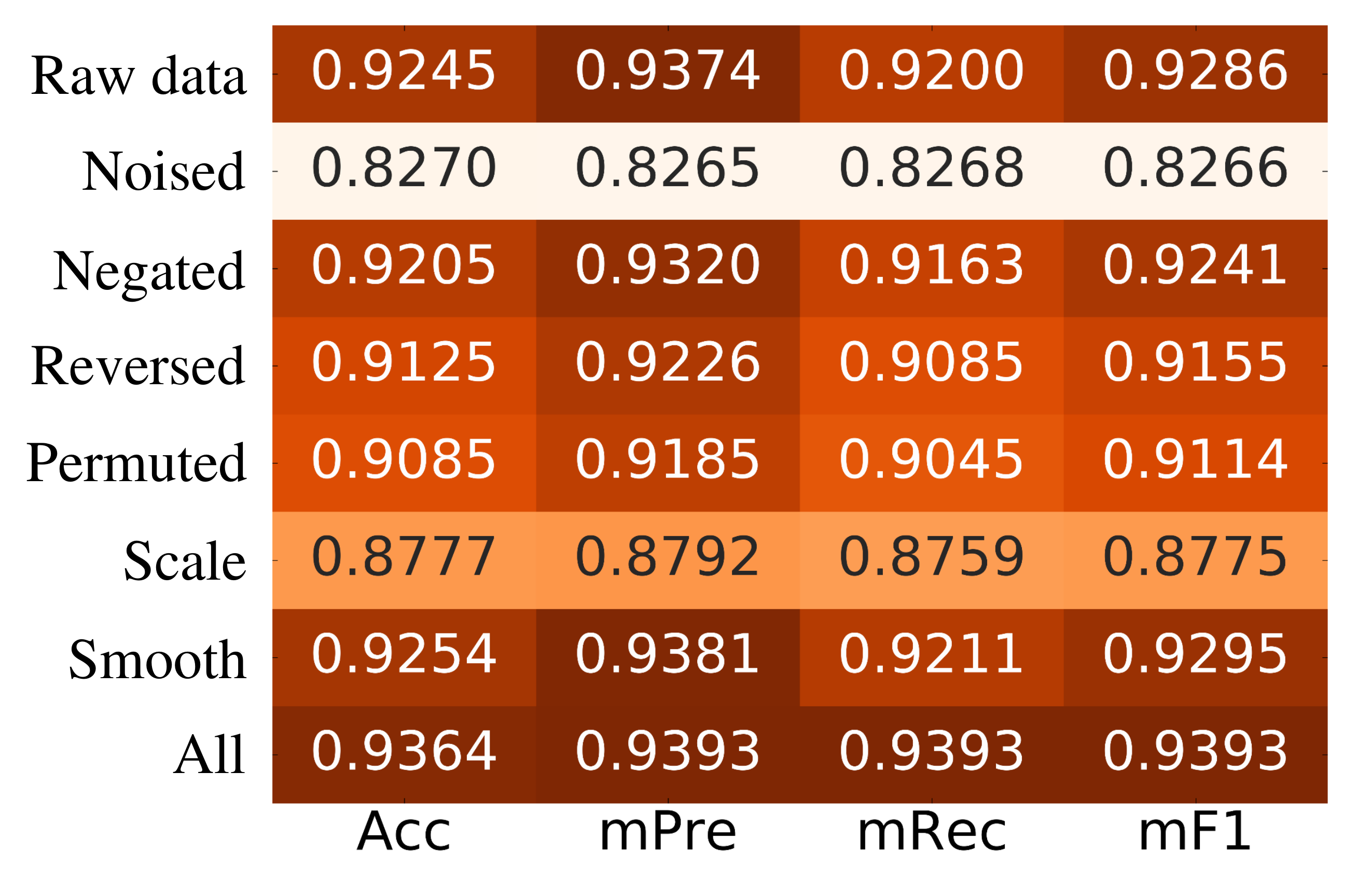}
    \label{fig-sub-ssl}
    }
    \subfigure[Adaptive memory fusion]{
    \includegraphics[width=.3\textwidth]{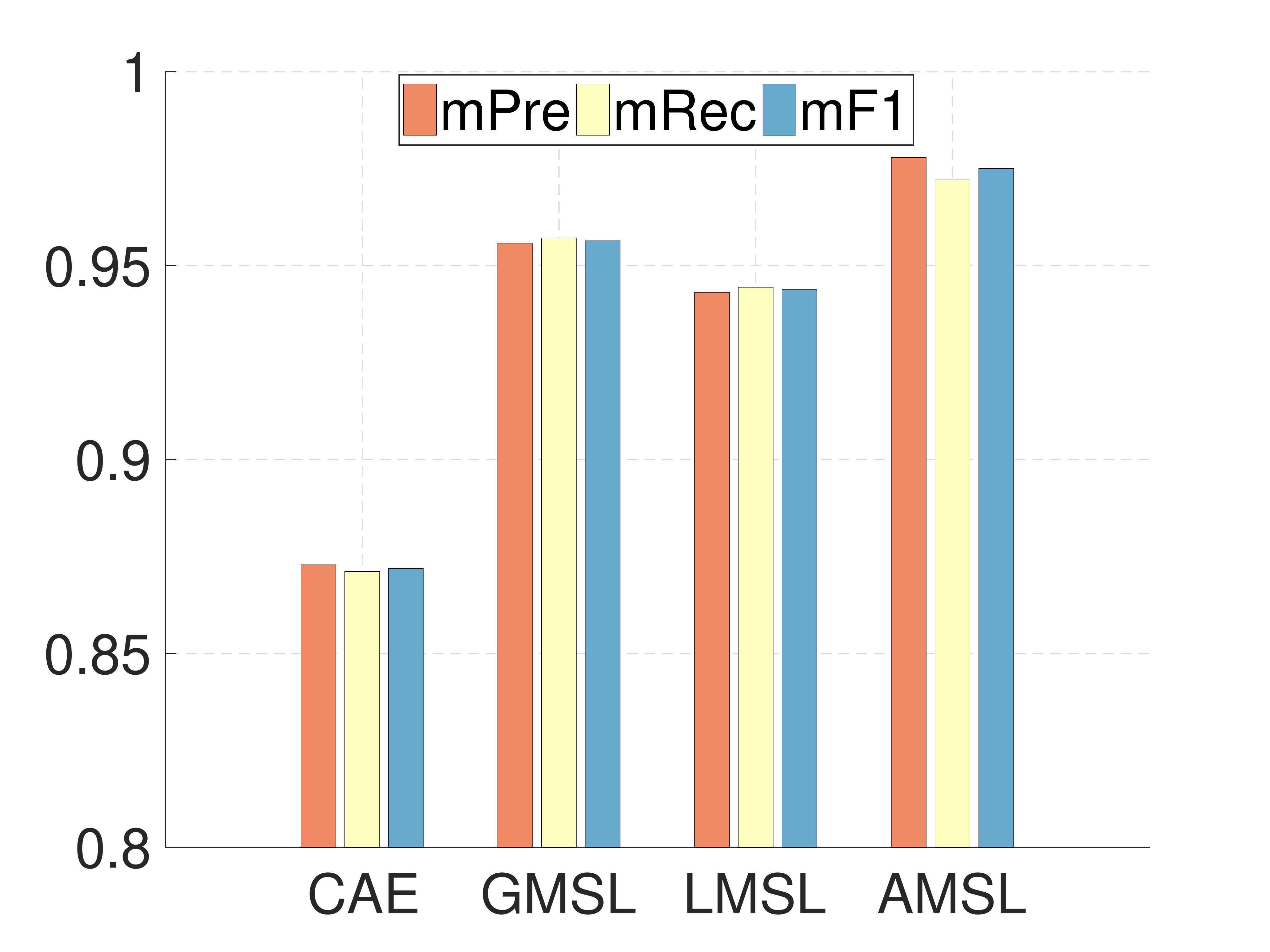}
    \label{fig-sub-memory}
    }
    \subfigure[Learned weights]{
    \includegraphics[width=.3\textwidth]{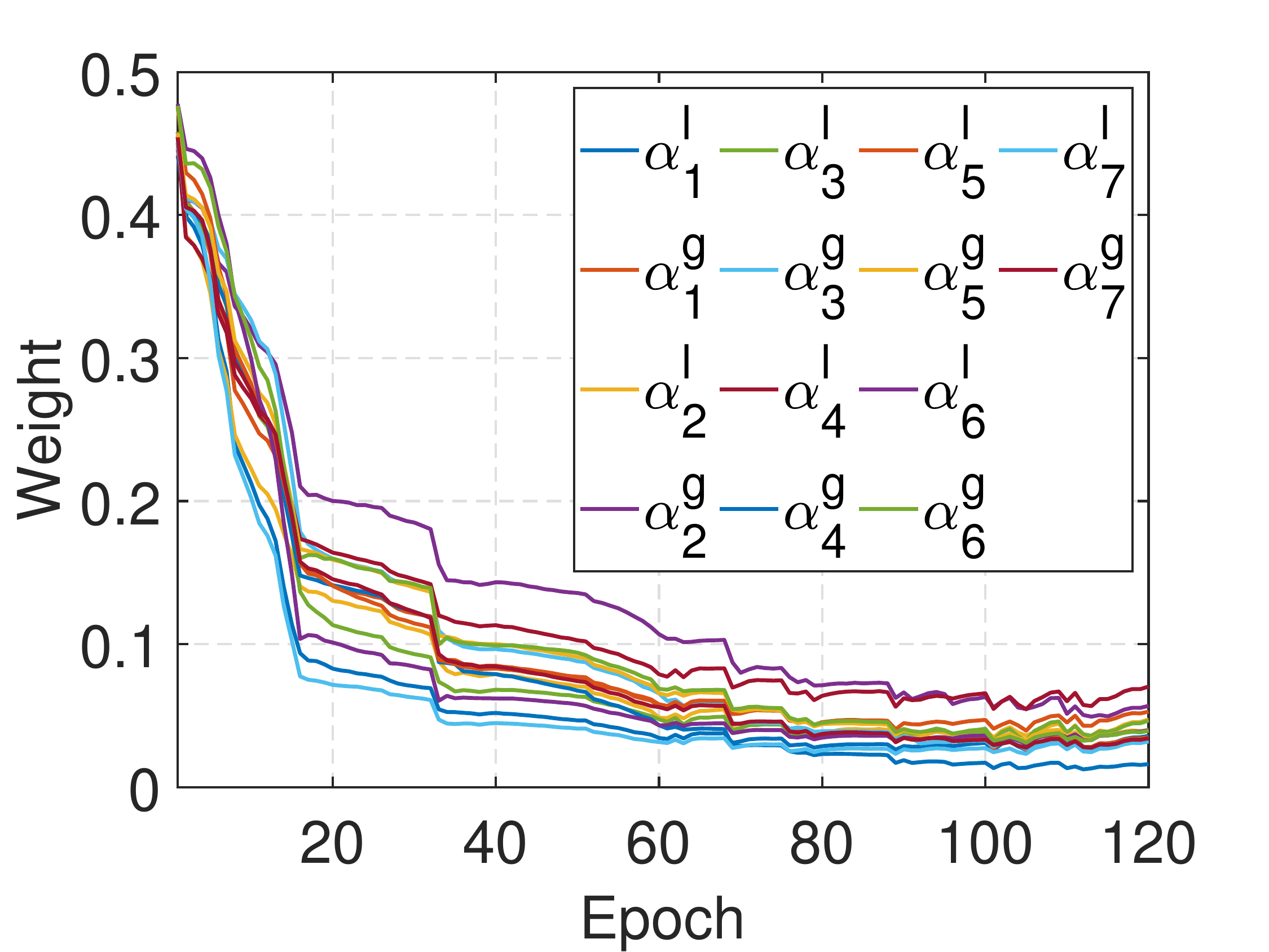}
    \label{fig-sub-weight}
    }
    \caption{Analysis of \method from several key aspects on PAMAP2 dataset. (a) The effectiveness of each data transformation in self-supervised learning. (b) Comparison between local, global, and our adaptive memory fusion modules. (c) The weights learned by our adaptive memory fusion module.}
    \label{fig:ana}
\end{figure*}

Traditional methods perform differently on different datasets since they are limited by the feature extraction methods and the curse of dimensionality. In deep learning methods, CNN-LSTM achieves the lowest F1 score, which means CNN-LSTM alone is not enough to capture general patterns and more regularizations are needed to improve its performance. For the reconstruction-based methods (LSTM-AE, MSCRED, and ConvLSTM-AE), their performances are limited by the noise in the training data, which is likely to mislead the model by making it hard to distinguish between the normal, abnormal, and noisy data.
MNAD and ConvLSTM are proposed for video data, which may not be suitable for multivariate time series. For BeatGAN, it has poor performance on CAP and WESAD datasets since it is prone to mode collapse and convergence problems on large-scale datasets. UODA performs reasonably well on the PAMAP2 and DSADS datasets that relies on pre-training denoising autoencoder (DAEs) and deep recurrent networks (RNNs) before fine-tuning. In fact, its performance is still not optimal because RNN does not have the capability to memorize long-term time series. GDN has a good performance on PAMAP2 and DSADS datasets. However, with the increase of data, the running speed of the model slows down and the accuracy also decreases affected by the graph structure. For ConvLSTM-COMPOSITE model, it performs better than most baselines. However, its efficiency may be limited due to the existence of two decoders in its structure. 

\begin{figure}[t!]
    \centering
    \subfigure[DSADS]{\includegraphics[width=.2\textwidth]{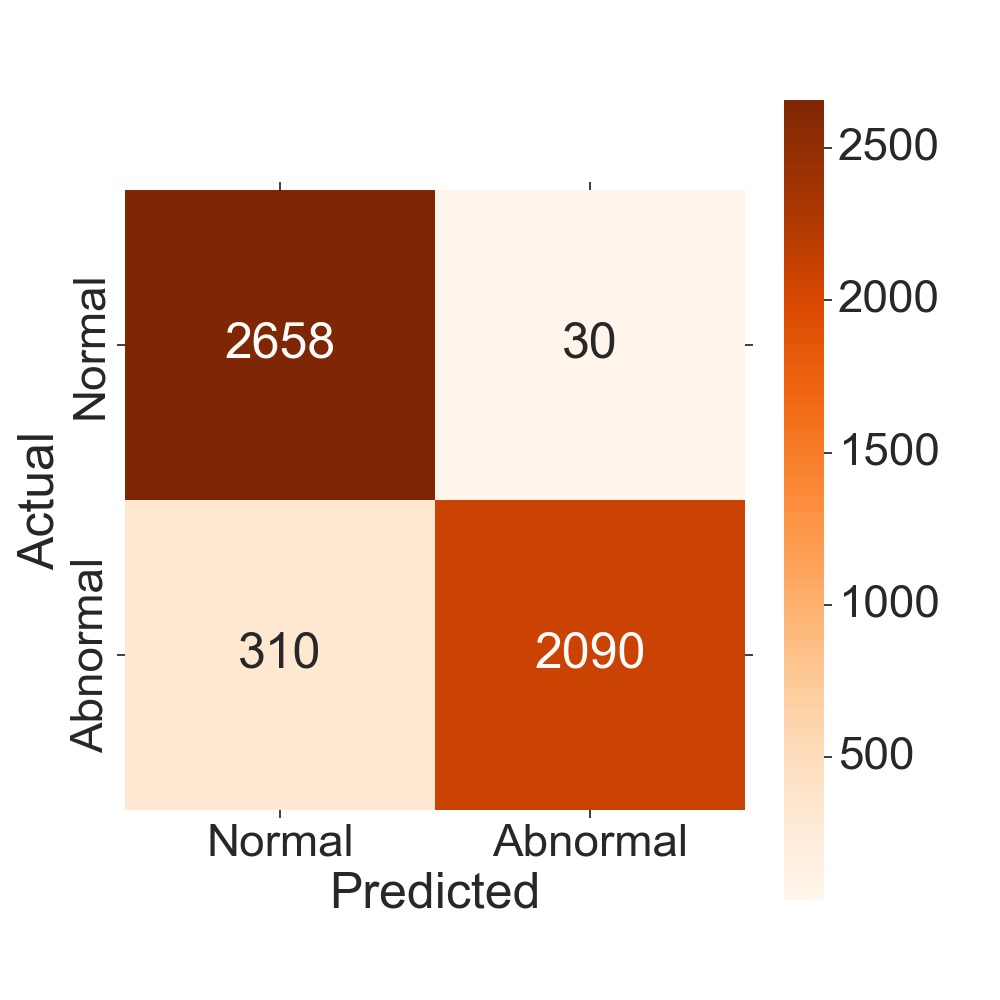}
    \label{fig-confusion1}}
    \subfigure[PAMAP2]{\includegraphics[width=.2\textwidth]{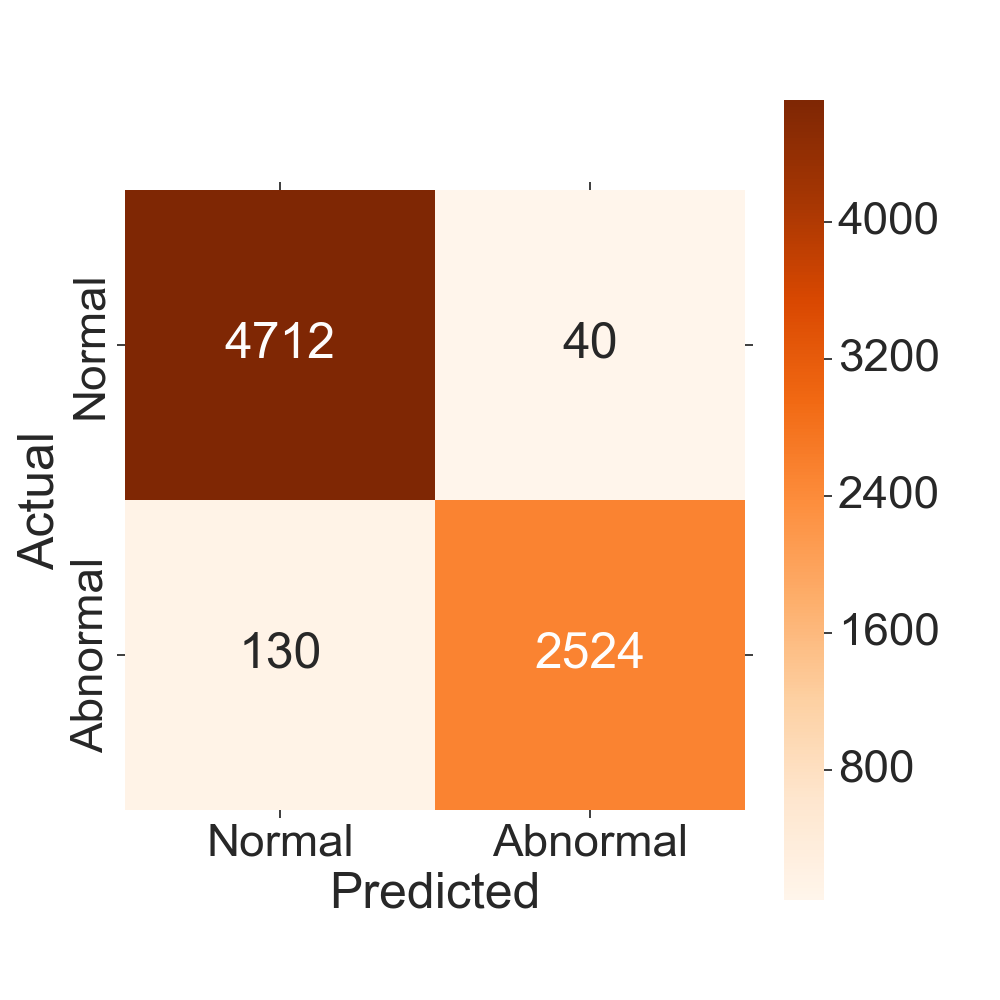}
    \label{fig-confusion2}}
    \subfigure[WESAD]{
    \includegraphics[width=.2\textwidth]{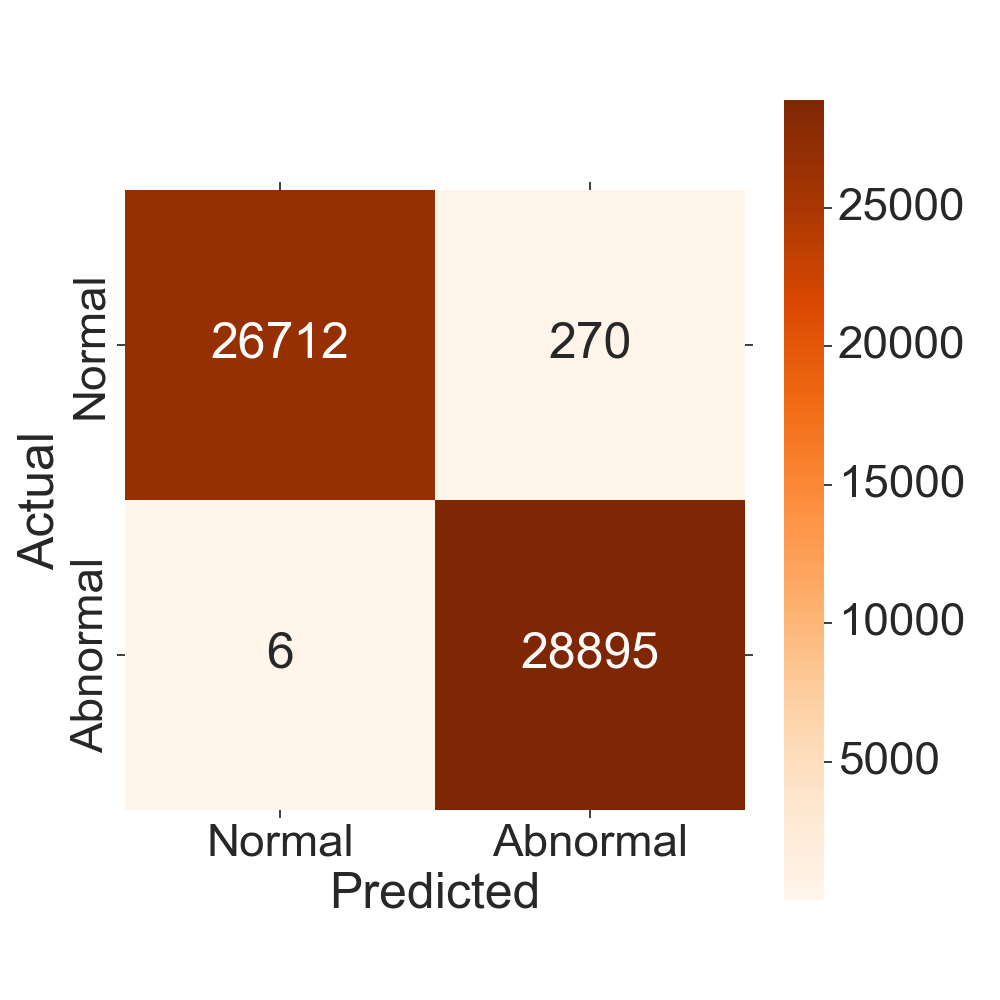}
    \label{fig-confusion3}
    }
    \subfigure[CAP]{
    \includegraphics[width=.2\textwidth]{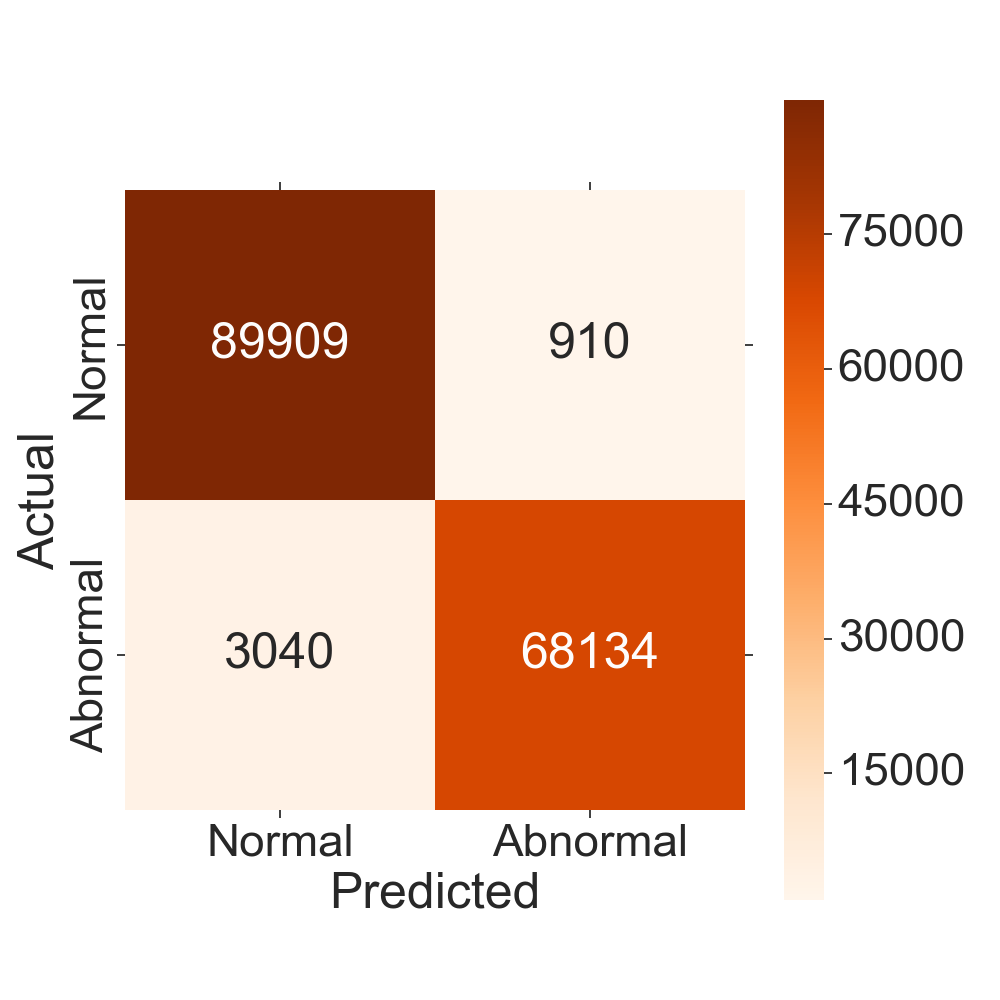}
    \label{fig-confusion4}
    }
    \caption{The confusion matrices on four datasets.}
    \label{fig:confusion}
\end{figure}

Besides, we make confusion matrices to error analysis for our proposed method AMSL. As shown in \figurename~\ref{fig:confusion}, we find that the proportion of misclassification of normal data is lower than that of abnormal data for most datasets. The F1-scores of normal class and abnormal class are 93.99\% and 92.48\% on DSADS dataset, the F1-scores of normal class and abnormal class are 98.23\% and 96.74\% on PAMAP2 dataset, the F1-scores of normal class and abnormal class are 99.49\% and 99.52\% on WESAD dataset, the F1-scores of normal class and abnormal class are 97.85\% and 97.18\% on CAP dataset. The proposed AMSL method achieves significantly superior performance in all the datasets, with an F1 score of at least 93\%.

\subsection{Ablation Study}

In this section, we conduct ablation study to show the effectiveness of the self-supervised learning, memory and adaptive fusion modules of \method. All experiments were performed using the PAMAP2 dataset. \tablename~\ref{tb-ablation1} reports the performance of different variants of \method. These results show that the proposed self-supervised learning module and memory module can both increase the performance of the base convolutional autoencoder (CAE) module. After combining these two modules, it can be observed that \method can achieve better performance than that without the adaptive fusion module. Here, ``Mem'' denotes memory fusion module. This means that $z^g_i$ and $z^l_i$ are fused in a proportion of $1:1$. 

\method also consists of other components, such as self-supervised loss and sparse loss. To demonstrate the effectiveness of these components, we conduct other ablation studies shown in \tablename \ref{tb-ablation1} (last two rows). It is worth noting that CAE + SSL without self-supervised loss (i.e., only using different augmentations) obtain a lower F1 score than other variants of the approach. This is because the augmented data (via different transformations) complicates the distribution of normal data, thus leading to the underfitting problem. Therefore, it is necessary to classify these transformations using pseudo labels. Similarly, we observe that adding sparse loss further improves the performance. 
\begin{table}[t!]
    \caption{Ablation results on the PAMAP2 dataset. ``SSL'' denotes self-supervised learning module.}
    \label{tb-ablation1}
    \begin{tabular}{ccccc}
    \toprule
        Variant & mPre & mRec & mF1 & Acc \\ \hline
        CAE & 0.8728 & 0.8711 & 0.8719 & 0.8676 \\
        CAE + Mem & 0.8808 & 0.8819 & 0.8813 & 0.8809 \\
        CAE + SSL &0.9393  &0.9393  &0.9393  &0.9364  \\
        CAE + SSL + Mem   & 0.9616  & 0.9420   & 0.9517 & 0.9554\\
        CAE + SSL + Ada Mem & \textbf{0.9788} & \textbf{0.9713} & \textbf{0.9750} & \textbf{0.9770} \\ 

        \hline
        CAE + Mem w/o \ Spar & 0.8713  & 0.8723   & 0.8719 & 0.8718   \\
        CAE + SSL w/o SSL loss   & 0.8231  & 0.8175   & 0.8203 & 0.8205   \\
    \bottomrule
    \end{tabular}
\end{table}

\subsection{Detailed Analysis}

\subsubsection{Self-Supervised Learning}
Self-supervised learning helps the network learn general and diverse features for the normal data, thus increasing the generalization ability of the model to recognize the unseen normal and abnormal instances.
In \figurename~\ref{fig-sub-ssl}, we show comparative performance analysis of each self-supervised data transformation. This assessment helps us understand whether model performance via jointly learning augmented data is better than learning individual data. The experiments were performed using the PAMAP2 dataset. The results show that the overall performance is competitive except for the noisy signals, thus it is beneficial to combine all transformations for better generalization. In Section \ref{conver}, we discard the poorly performing transformations such as ``Noise'' and ``Scale''. We observe that the F1 score and accuracy of AMSL decrease as transformations $R$ decrease. This demonstrate that it's helpful to use different self-supervised data transformations to increase the model’s generalization ability.

\begin{figure*}[t!]
    \centering
    \includegraphics[width = 1\textwidth]{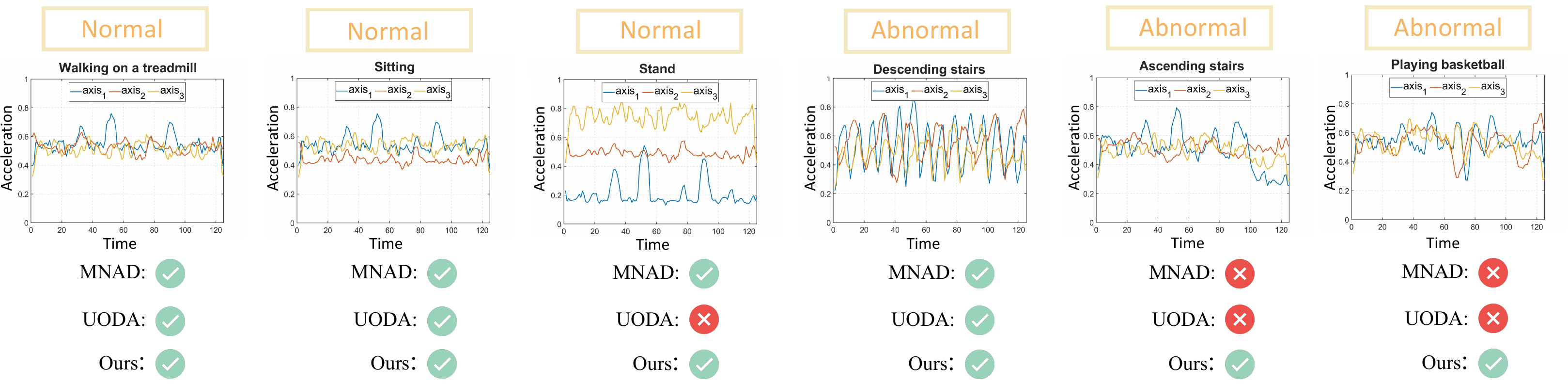}
    \caption{The visualization for samples in normal and abnormal class. Each column is the instance whether is correctly detected by our method \method, MNAD and UODA. $Axis_{1,2,3}$ represents three dimensional signals on DSADS dataset.}
    \label{fig:case}
\end{figure*}

\subsubsection{Adaptive Memory Fusion module}
In \figurename~\ref{fig-sub-memory}, we show the performance of CAE, GMSL, LMSL and \method.
GMSL and LMSL are global memory network and local memory network with SSL. The experimental results demonstrate that adaptive memory fusion network achieves better performance than using single memory network (i.e. global and local memory module). \method achieves a $\textbf{1.86\%}$ higher F1-score than GMSL, and a $\textbf{3.12\%}$ higher F1-score than LMSL. The same goes for the precision and recall. More results on different datasets are shown in \tablename~\ref{tb-ablation3}. Our method \method is always superior than GMSL and LMSL.
As mentioned above, \method finds the best weight $\alpha^l$ and $\alpha^g$ values automatically. To observe the changes of adaptive weights at the training stage, this experiment was conducted using the PAMAP2 dataset. As shown in \figurename~\ref{fig-sub-weight}, it occurred at around the 70-th epoch when the values of adaptive weights stabilized. Here, $\alpha_1-\alpha_7$ denote the weights of raw data and 6 transformations corresponding to transformations in \figurename~\ref{fig-sub-ssl}.

\begin{table}[htbp]
    \caption{Ablation results on all these datasets.}
    \centering
    \label{tb-ablation3}
    \subtable[DSADS]{
    \begin{tabular}{ccccc}
    \toprule
        Variant & mPre & mRec & mF1 & Acc \\ \hline
        GMSL & 0.8890 & 0.8551 & 0.8717 & 0.8624 \\
        LMSL & 0.8915  &0.8523  & 0.8716 & 0.8605 \\
        AMSL & \textbf{0.9407} & \textbf{0.9298} & \textbf{0.9352} & \textbf{0.9332} \\ \bottomrule
    \end{tabular}
    }
    \subtable[PAMAP2]{
    \begin{tabular}{ccccc}
    \toprule
        Variant & mPre & mRec & mF1 & Acc \\ \hline
        GMSL & 0.9558 & 0.9571 & 0.9564 & 0.9553 \\
        LMSL & 0.9431  &0.9444  & 0.9438 & 0.9433 \\
        AMSL & \textbf{0.9788} & \textbf{0.9713} & \textbf{0.9750} & \textbf{0.9770} \\ \bottomrule
    \end{tabular}
    }
    \subtable[WESAD]{
    \begin{tabular}{ccccc}
    \toprule
        Variant & mPre & mRec & mF1 & Acc \\ \hline
        GMSL & 0.9863 & 0.9849 & 0.9856 & 0.9851 \\
        LMSL & 0.9864 & 0.9850 & 0.9857 & 0.9855 \\
        AMSL & \textbf{0.9953} & \textbf{0.9949} & \textbf{0.9951} & \textbf{0.9951} \\ \bottomrule
    \end{tabular}
    }
    \subtable[CAP]{
    \begin{tabular}{ccccc}
    \toprule
        Variant & mPre & mRec & mF1 & Acc \\ \hline
        GMSL & 0.9638 & 0.9556 & 0.9597 & 0.9598 \\
        LMSL & 0.9620  &0.9531  & 0.9575 & 0.9575 \\
        AMSL & \textbf{0.9771} & \textbf{0.9736} & \textbf{0.9753} & \textbf{0.9756} \\ \bottomrule
    \end{tabular}
    }
\end{table}

\subsection{Robustness to Noisy Data}
\begin{figure}[htbp]
    \centering
    \subfigure[Precision]{\includegraphics[width=.15\textwidth]{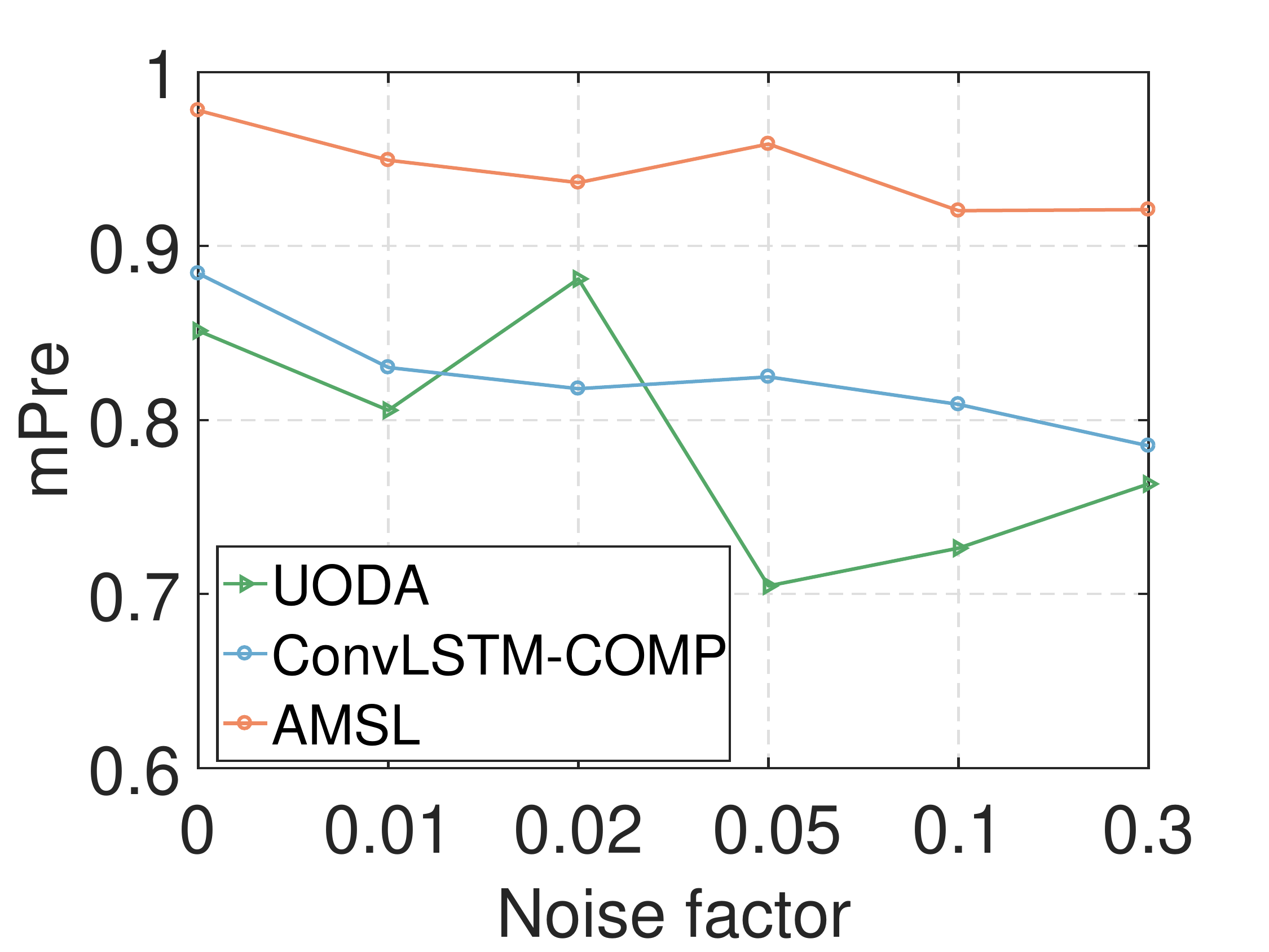}
    \label{fig-sub-noise1}}
    \subfigure[Recall]{\includegraphics[width=.15\textwidth]{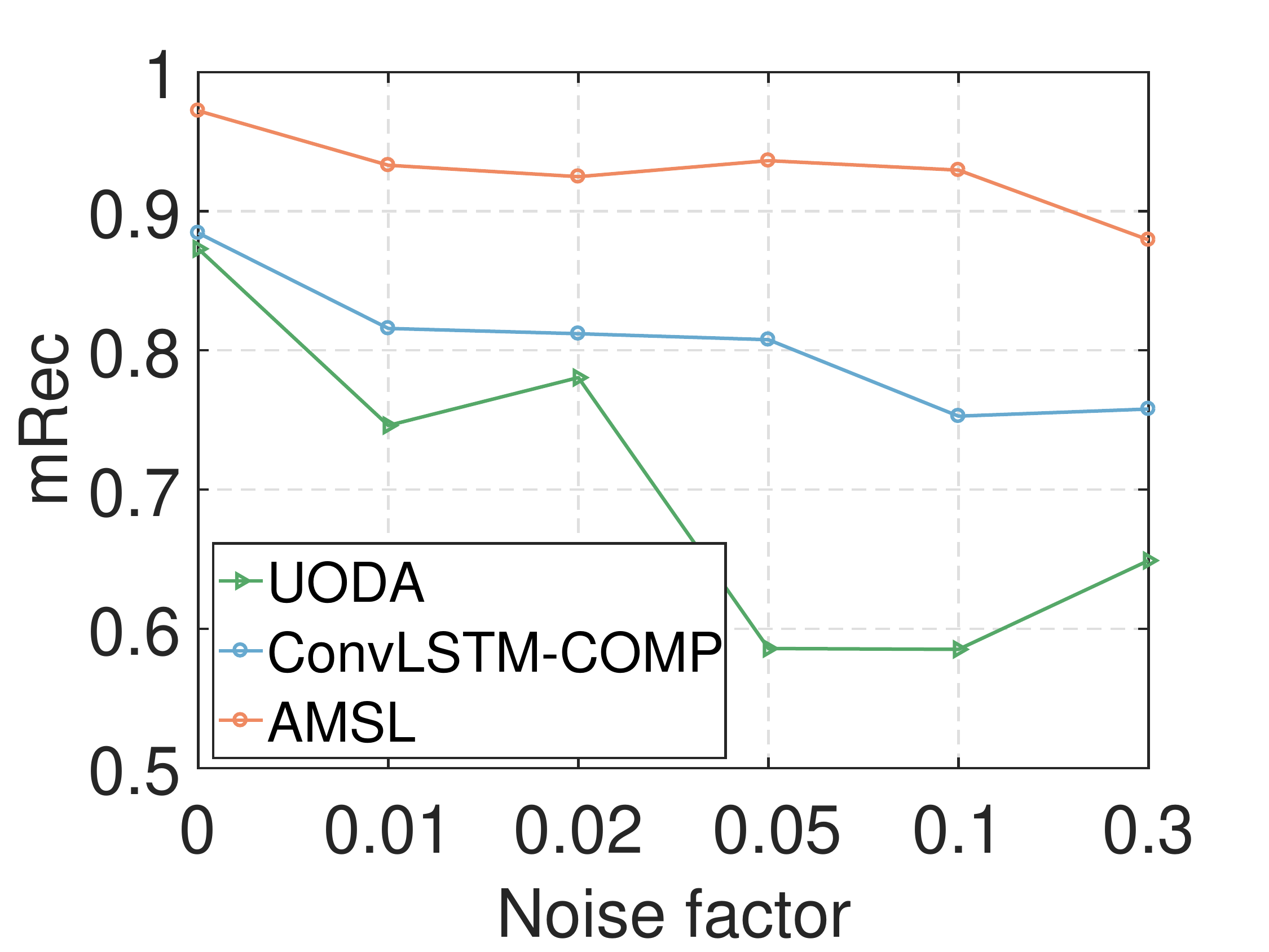}
    \label{fig-sub-noise2}}
    \subfigure[F1]{
    \includegraphics[width=.15\textwidth]{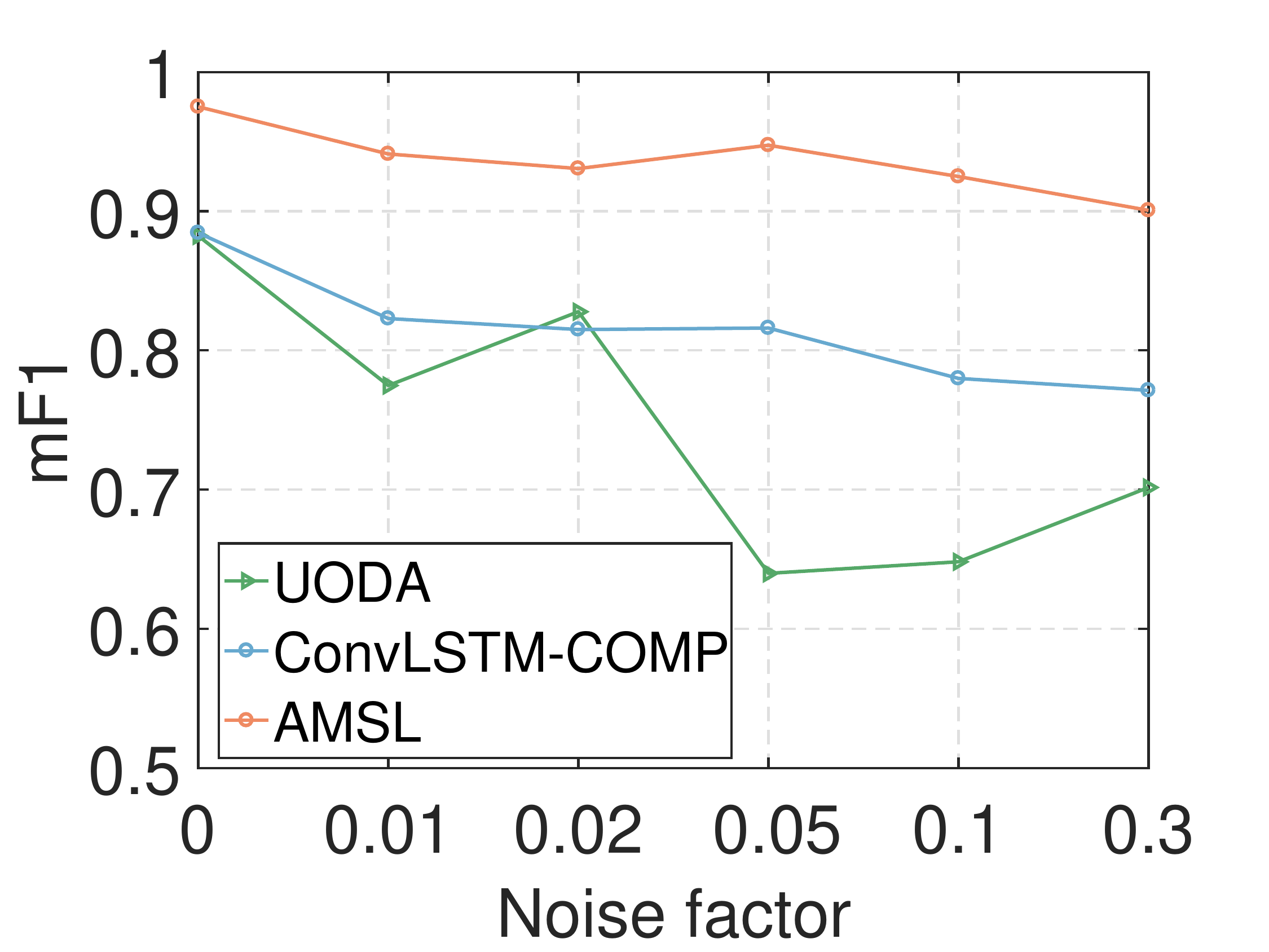}
    \label{fig-sub-noise}
    }
    \caption{Robustness to noise.}
    \label{fig:noise}
\end{figure}

In real-world applications, the collection of multivariate time series data can be easily polluted with noise due to changes in the environment or the data collection devices.
The noisy data bring critical challenges to the unsupervised anomaly detection methods.
In this section, we evaluate the robustness of different methods to noisy data on PAMAP2 dataset.
We manually control the noisy data ratio in the training data. We inject Gaussian noise ($\mu =0, \sigma = 0.3$) in a random selection of samples with a ratio varying between 1\% to 30\%. We compare the performance of three methods: UODA, ConvLSTM-Composite, and \method in \figurename~\ref{fig-sub-noise}. 
As the noise increases, the performance of all methods decreases.
Among them, \method remains significantly superior to others, demonstrating its robustness to noisy data.
The result of precision and recall are presented in \figurename~\ref{fig-sub-noise1} and \figurename~\ref{fig-sub-noise2}.

\subsection{Percentage of Anomaly}
\begin{figure}[htbp]
    \centering
    \includegraphics[width = 0.32\textwidth]{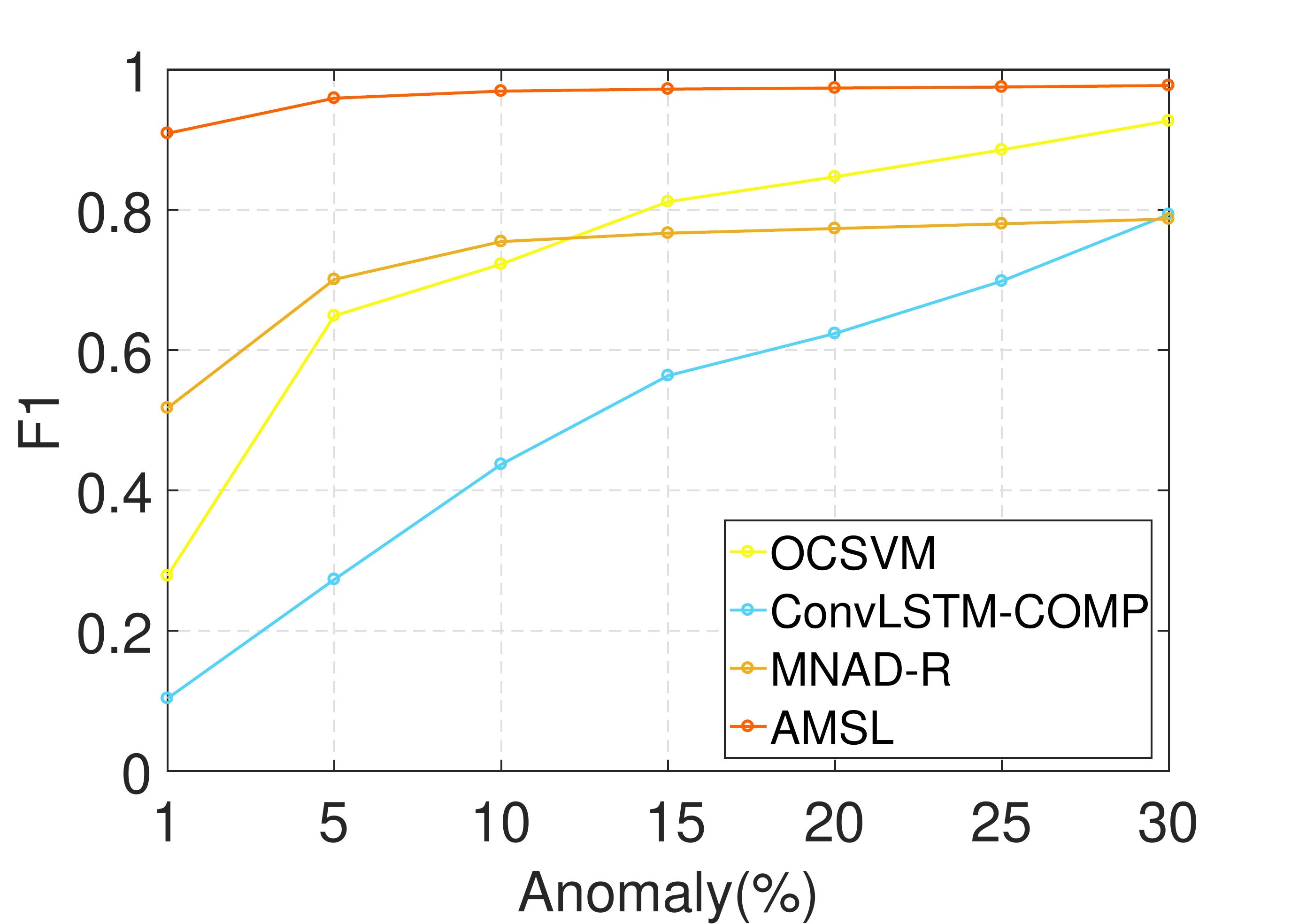}
    \caption{F1 score on the abnormal class with different percentage of anomaly.}
    \label{fig:anomaly}
\end{figure}
Generally, the percentage of anomaly is significantly lower than the normal range. Hence, we conduct experiments on CAP dataset when the percentage of anomaly on the testing set is 1\%, 5\%, 10\%, 15\%, 20\%, 25\% and 30\%. In \figurename~\ref{fig:anomaly}, we show the F1 score of the abnormal class using different methods. We compare the performance of four methods: OCSVM, ConvLSTM-COMPOSITE, MNAD-R and \method. These methods have good performance for CAP dataset in the above experiments. We can observe that as the percentage of anomaly decreases, the F1 score of other methods have decreased significantly, while our method still remain stable. This demonstrates that our method achieves high precision and recall on the abnormal class even if the percentage of anomaly is very low on the testing set. Therefore, we can come to a conclusion that our proposed method \method has good stability
even facing dataset imbalance problem.

\subsection{Case Study}
We present case studies of MNAD, UODA and our method \method as shown in \figurename~\ref{fig:case} by visualizing some normal and abnormal classes.
We choose three dimensional signals on DSADS dataset.
It can be shown that our \method can correctly classify these samples, while other methods fail in two situations: (1) when a normal sample is not similar to the majority of normal samples (i.e., overfitting) or (2) when an abnormal sample is very similar to the normal sample (i.e., less powerful representations).
This demonstrate that our \method can effectively handle the problem about the diversity of samples.

\subsection{Parameter sensitivity analysis}

\begin{figure}[htbp]
    \centering
    \subfigure[Window length $V$]{\includegraphics[width=.15\textwidth]{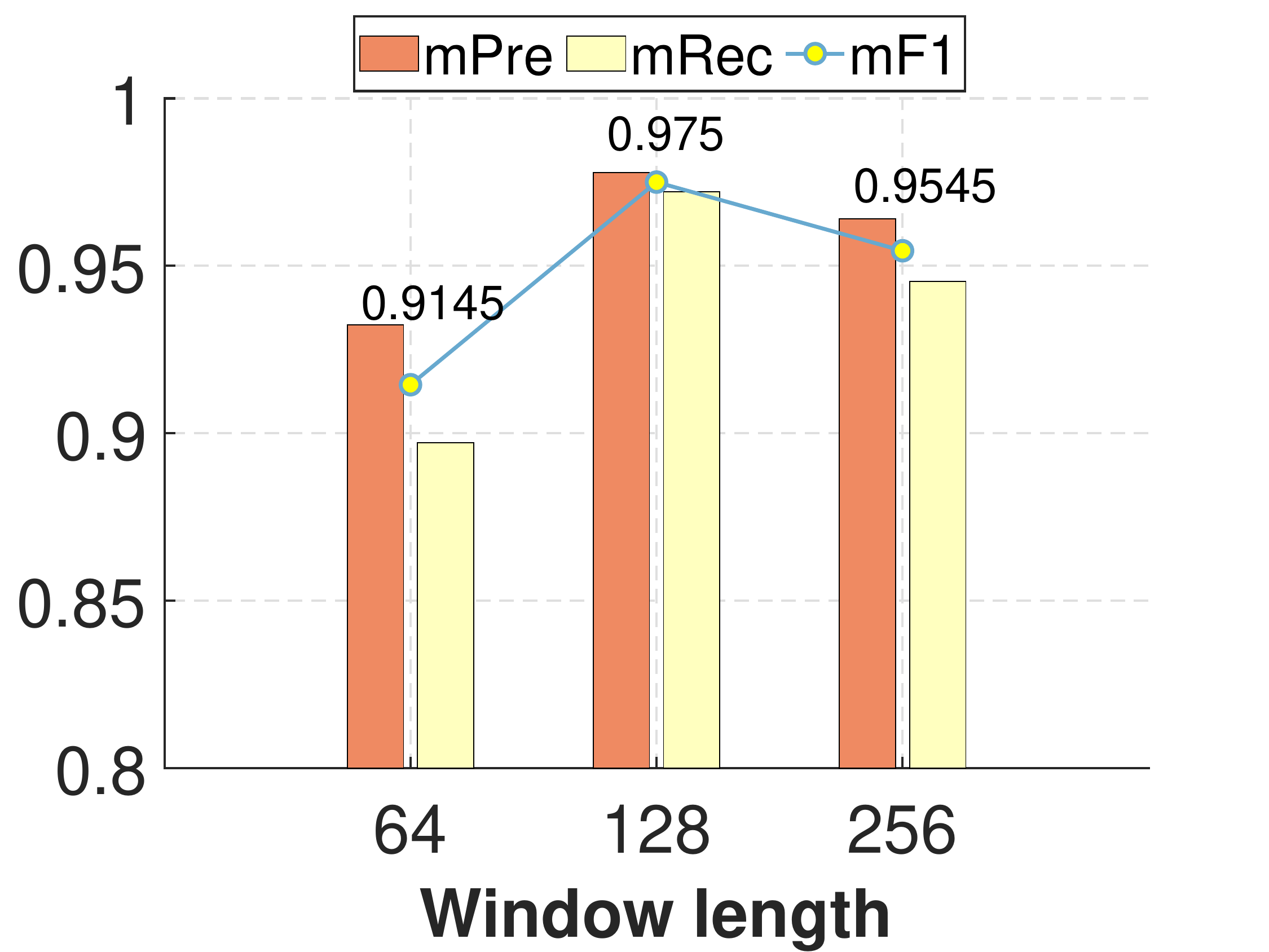}}
    \subfigure[Memory size $C$]{\includegraphics[width=.15\textwidth]{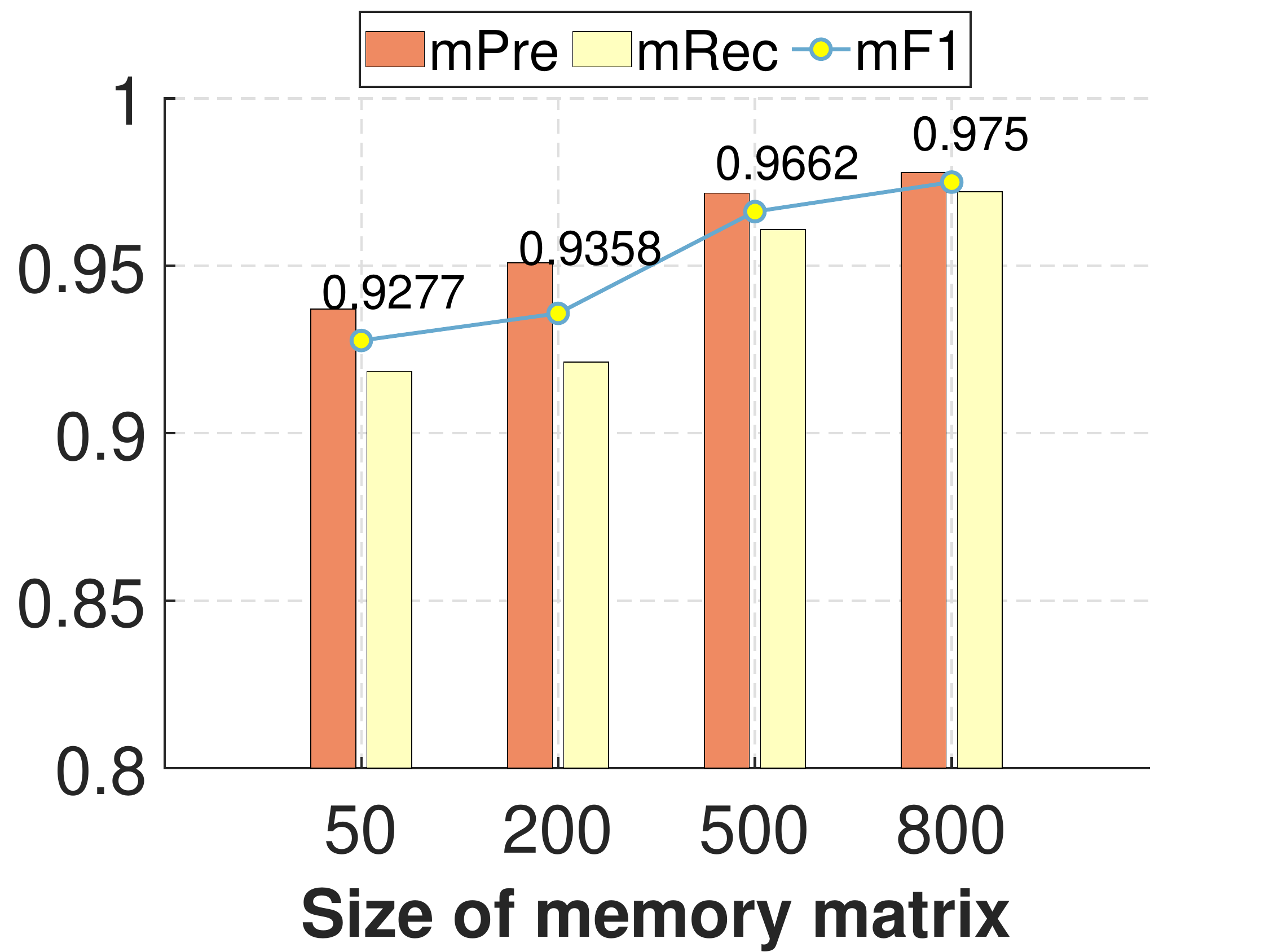}}
    \subfigure[Filter size $F$]{\includegraphics[width=.15\textwidth]{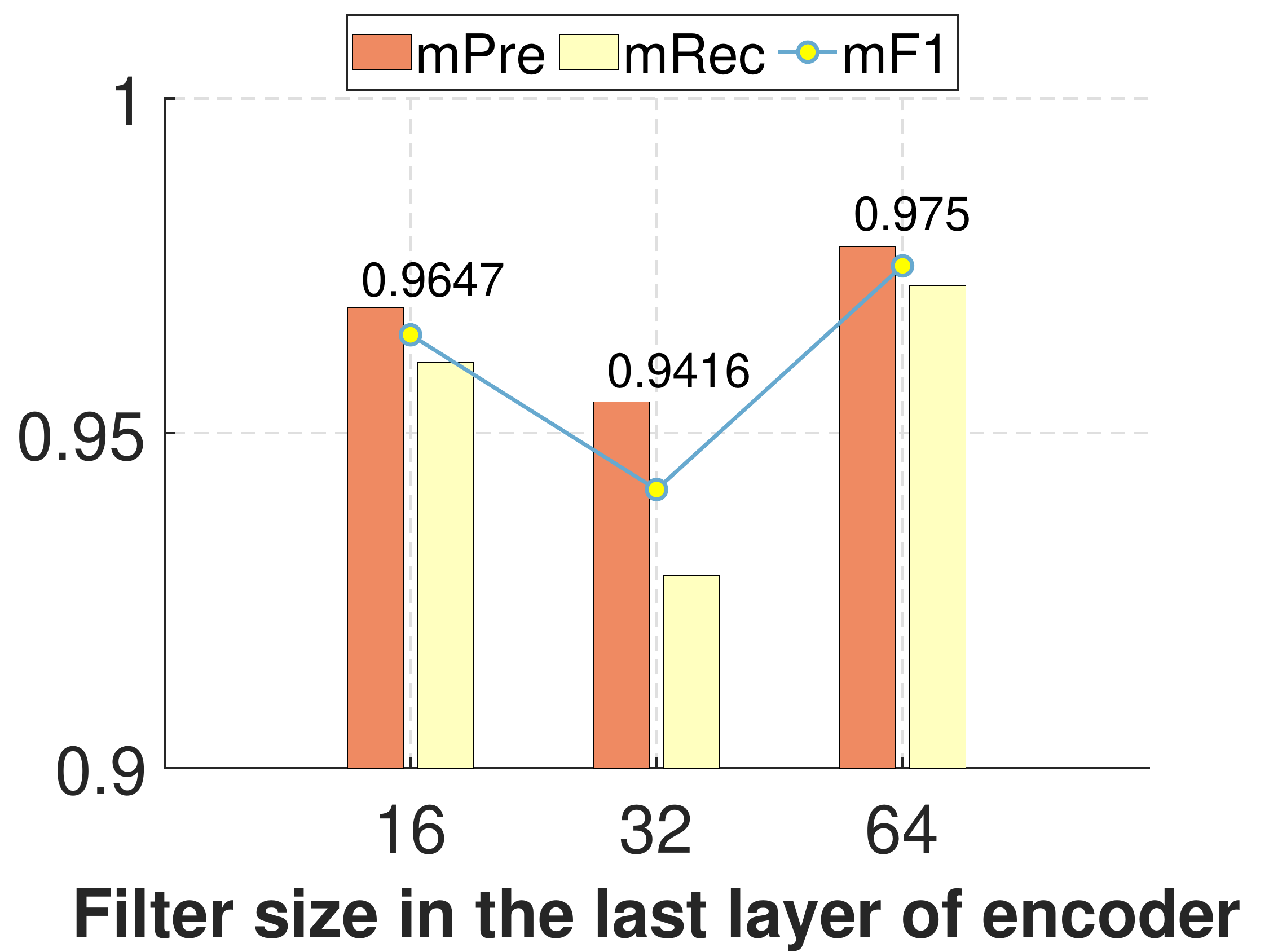}}
    \vspace{-.1in}
    \caption{Parameter sensitivity analysis on PAMAP2 dataset.}
    \label{fig-para-sens}
\end{figure}

We consider three key parameters: 1) window length $V$, 2) size of memory matrix $M$, and 3) filter size $F$ in the last layer of encoder.
These parameters are chosen from: $V \in \{64, 128, 236\}$, $C \in \{50, 200, 500, 800\}$, and $F \in \{16, 32, 64\}$.  \figurename~\ref{fig-para-sens} shows that \method is robust to different parameter choices on PAMAP2 dataset.

\begin{figure}[tp]
    \centering
    \subfigure[LMSL]{\includegraphics[width=.15\textwidth]{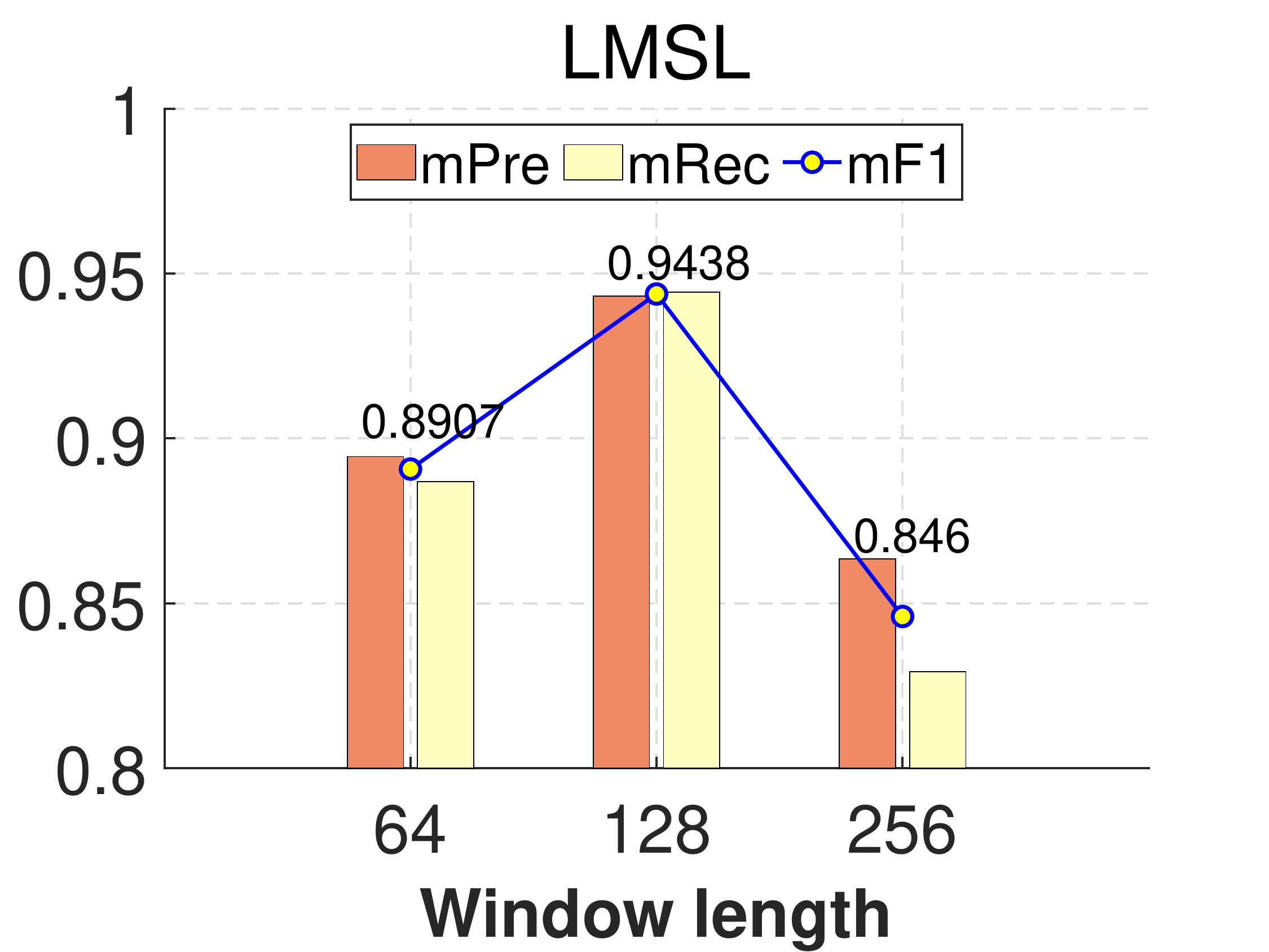}}
    \subfigure[GMSL]{\includegraphics[width=.15\textwidth]{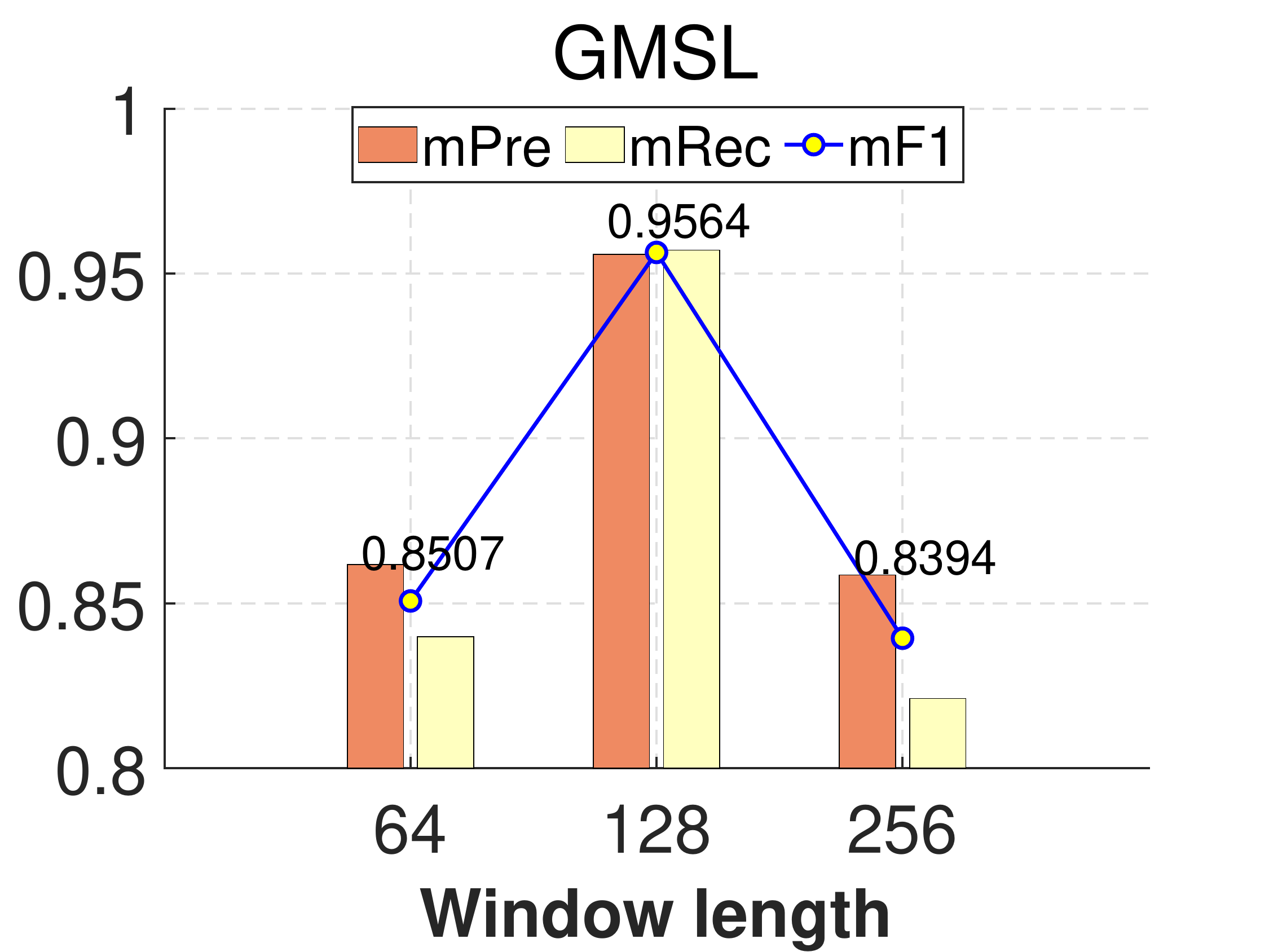}}
    \subfigure[LMSL]{\includegraphics[width=.15\textwidth]{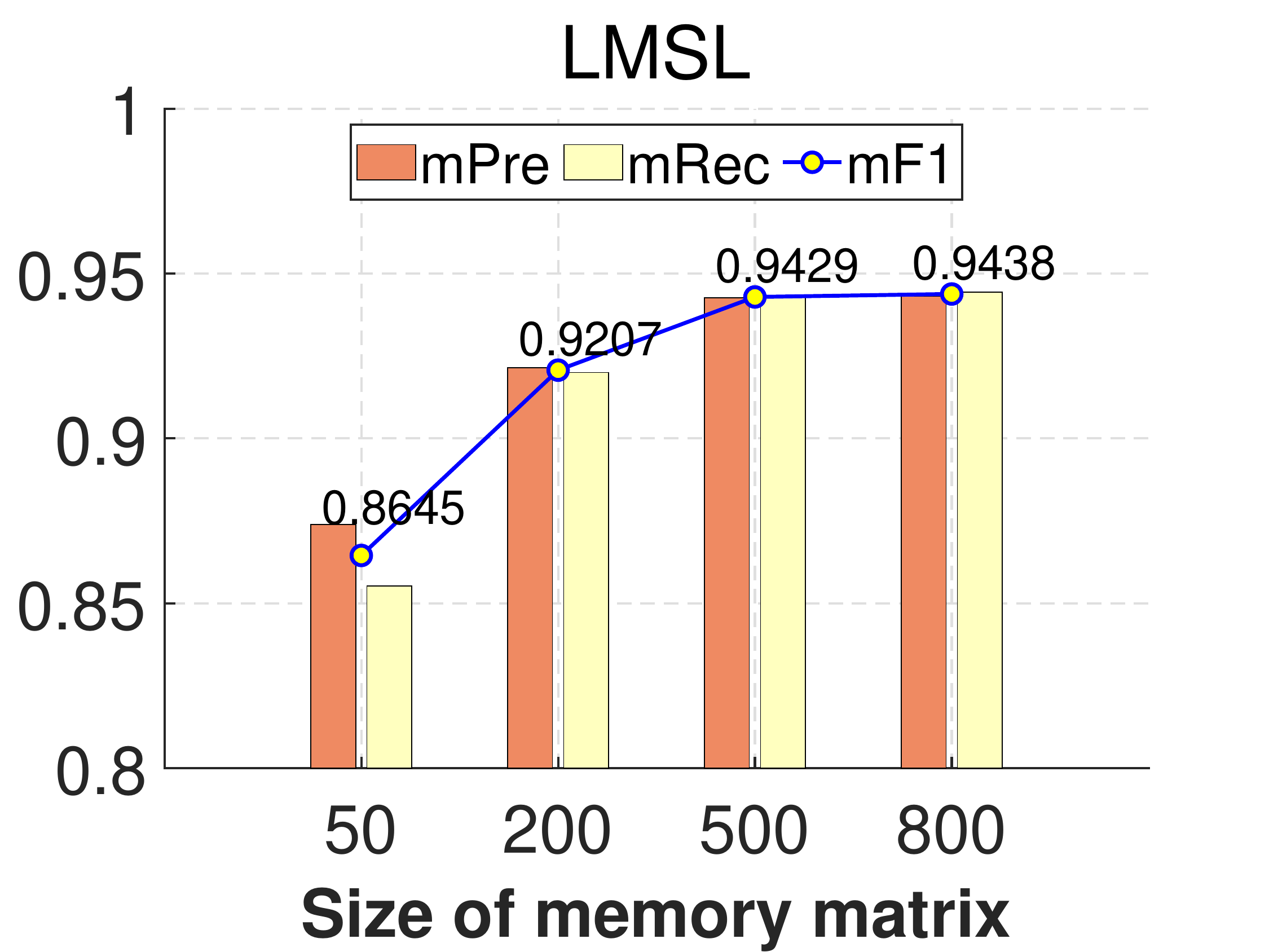}}
    \subfigure[GMSL]{\includegraphics[width=.15\textwidth]{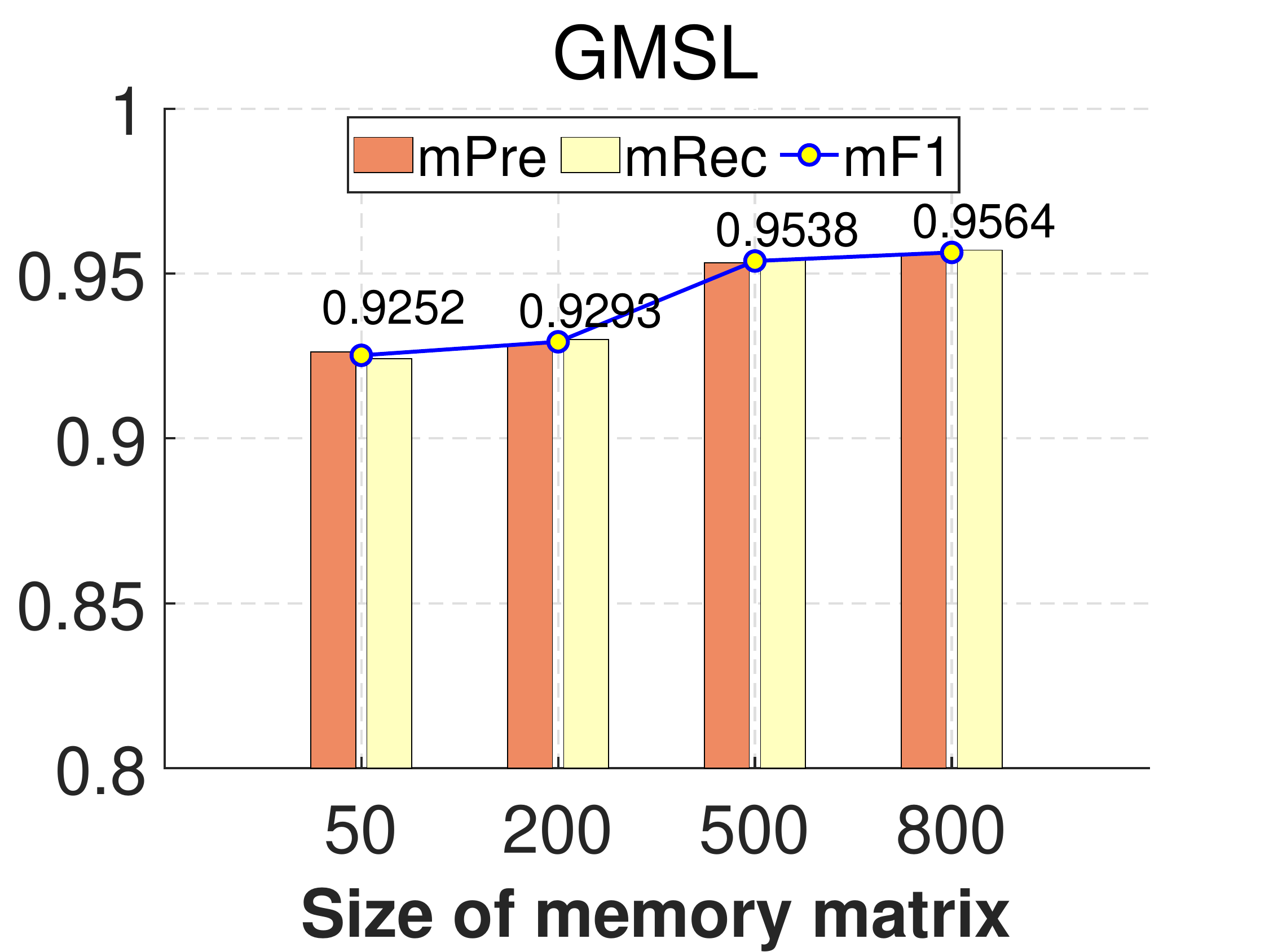}}
    \subfigure[LMSL]{\includegraphics[width=.15\textwidth]{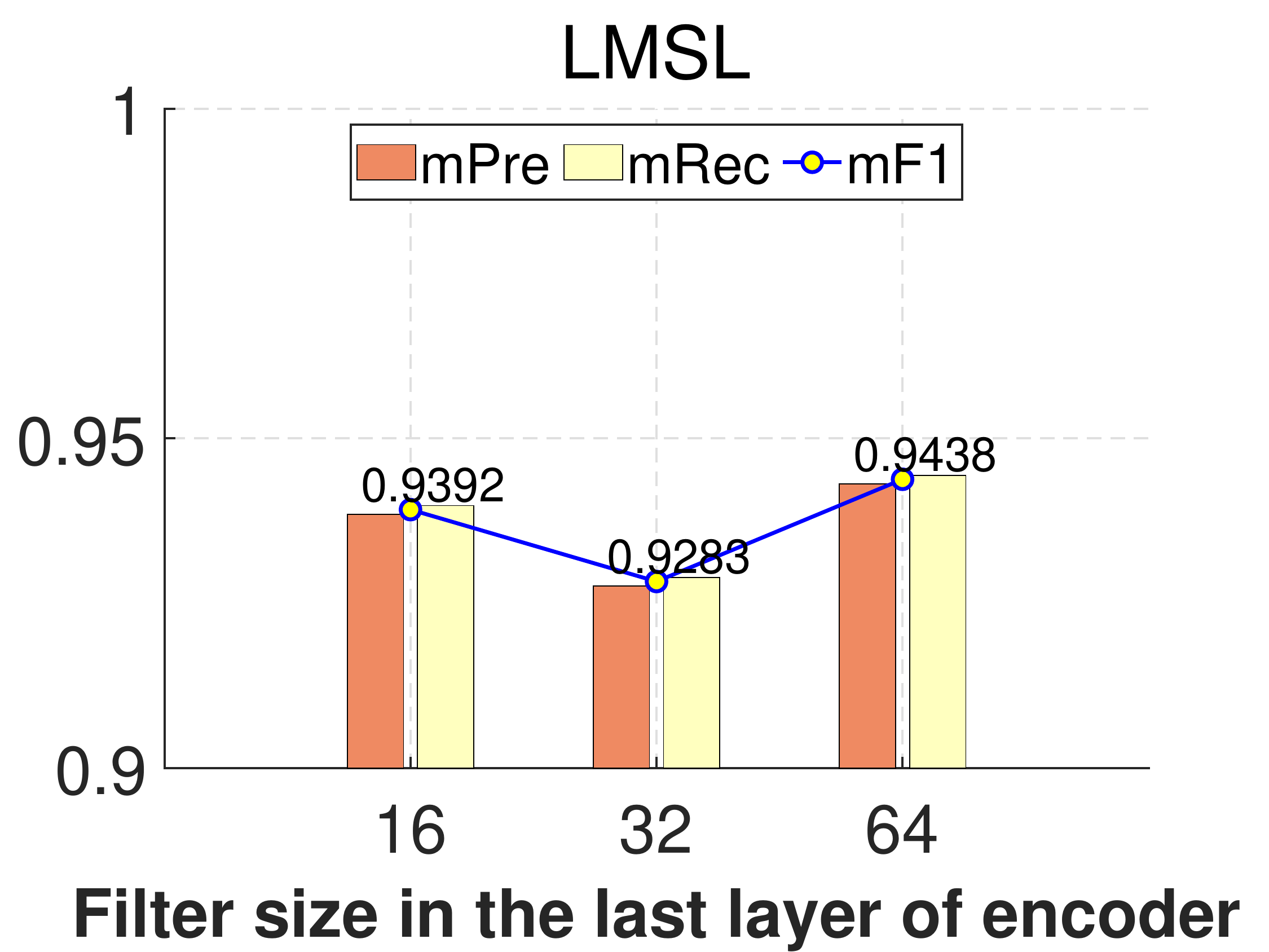}}
    \subfigure[GMSL]{\includegraphics[width=.15\textwidth]{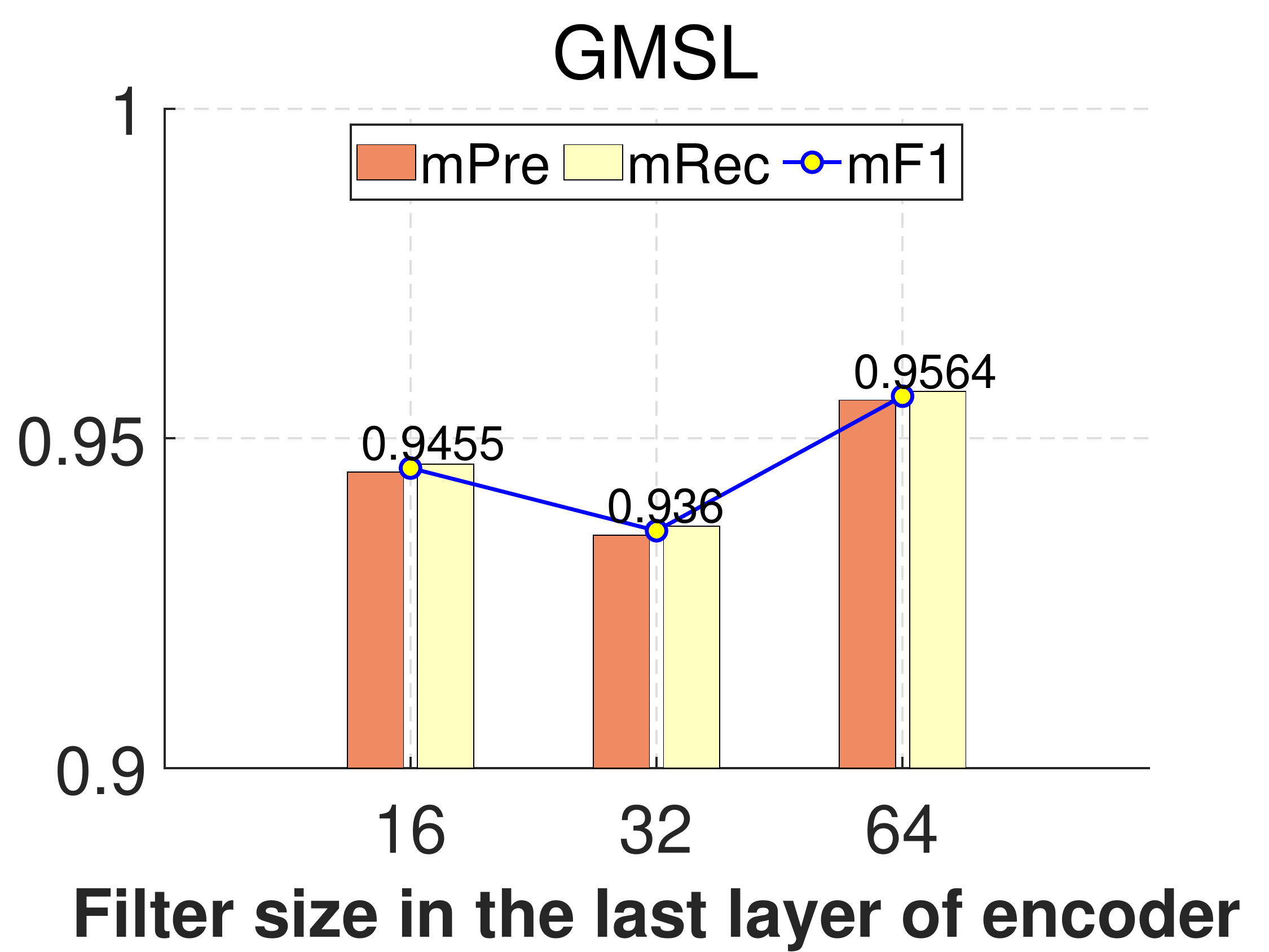}}
    \subfigure[$\lambda_1$]{\includegraphics[width=.15\textwidth]{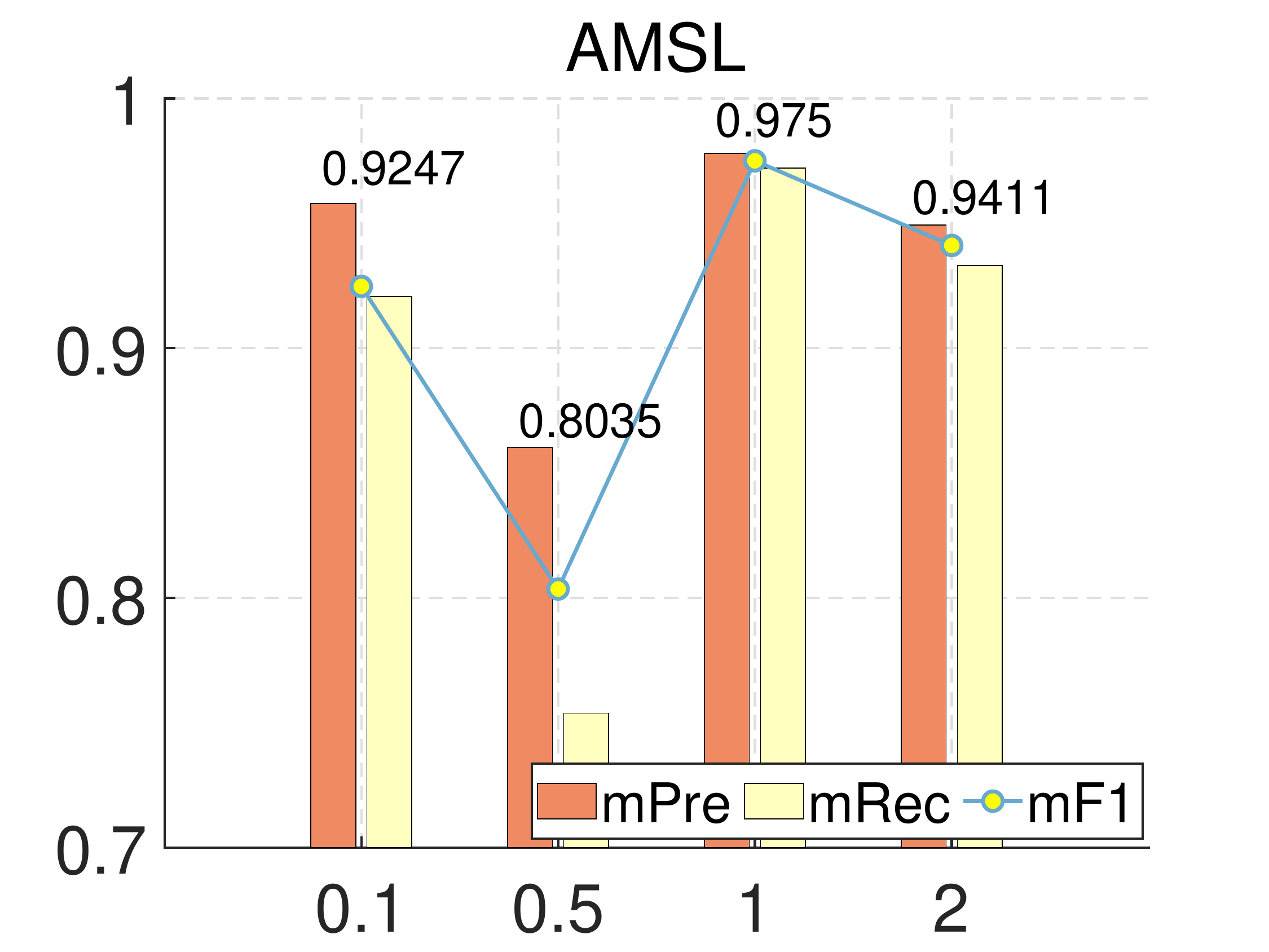}}
    \subfigure[$\lambda_2$]{\includegraphics[width=.15\textwidth]{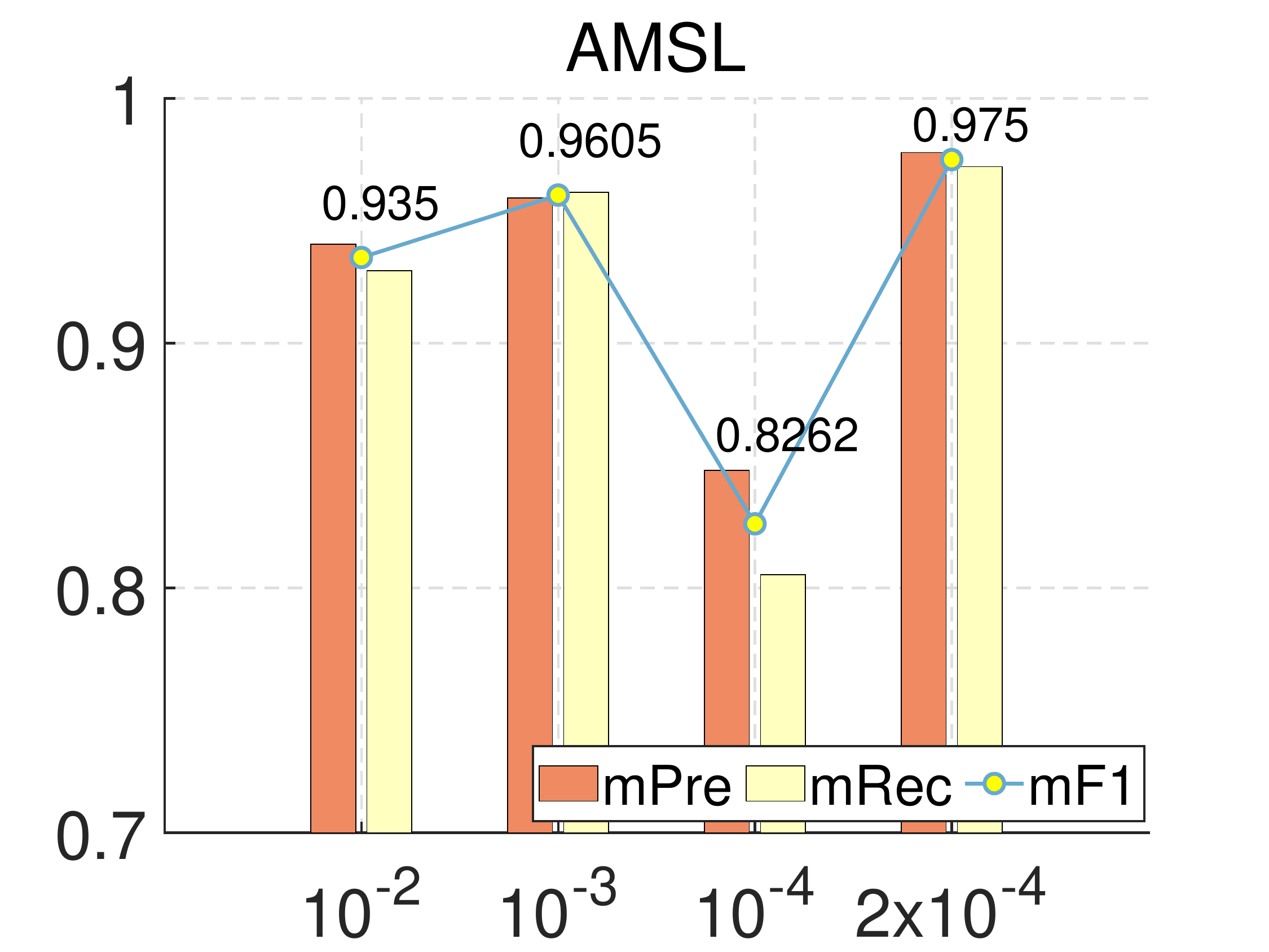}}
    \caption{Other parameter sensitivity analysis on PAMAP2 dataset.}
    \label{fig:con}
\end{figure}

We also provide the sensitivity analysis of LMSL and GMSL. \figurename~\ref{fig:con} (a-b) present the results of LMSL and GMSL with different window length $\{64, 128, 236 \}$.
We observe that the choice of window length is critical to the method, which window length of 128 can achieve the best performance for PAMAP2 dataset.
The second factor is the size of memory matrix $M \in \mathbb{R}^{ C \times F}$ that is set as $C \in \{50, 200, 500, 800\}$ and $F=64$. Note that the dimension of $F$ is equal to the filter size in the last layer of encoder.
As shown \figurename~\ref{fig:con} (c-d), they indicate that by increasing the size of $C$, the performance improves until $C$ reaches nearly 800.
\figurename~\ref{fig:con} (e-f) show the results for $F \in \{16, 32, 64\}$. The filter size in the last layer of encoder obtains the best performance when $F = 64$.
We use encoder to store more information and contribute it to reconstruct the latent features.

Moreover, we adjust the hyperparameters $\lambda_1$ and $\lambda_2$ in the loss function (Eq. ~\eqref{z19}).
As shown in \figurename~\ref{fig:con} (g-h), we observe that $\lambda_1=1$ and $\lambda_2=0.0002$ are the best choice in our experiments.

The selection of the threshold $\mu$ in Section \ref{threshold} is also an important issue for which we conduct experiments to compare different thresholds for $\alpha$-percentiles: 90, 95 and 99. As shown in \tablename~\ref{tab:thr}, we find that the 99th percentile can estimate the optimal threshold. Therefore, we set the anomaly detection threshold to the 99 percentile. 

\begin{table}[t!]
	\centering
	\caption{Experimental evaluation of threshold selection for AMSL.}  
	\label{tab:thr} 
	\begin{tabular}{lcccc}
		\toprule
		Decision Rules   & mPre & mRec & mF1 & Acc \\ 
		\midrule
		THR$_{90th \ percentile} $  & 0.9164  & 0.9401  & 0.9281  & 0.9284       \\
		THR$_{95th \ percentile} $  & 0.9486  & 0.9597  & 0.9541  & 0.9568        \\
		THR$_{99th \ percentile} $  & \textbf{0.9788} & \textbf{0.9713}  & \textbf{0.9750}     & \textbf{0.9770}  \\ \bottomrule
	\end{tabular}
\end{table}

\subsection{Convergence, Time and Space Complexity}
\label{conver}
\begin{figure}[t!]
    \centering
    \subfigure[Convergence]{
        \includegraphics[width=.22\textwidth]{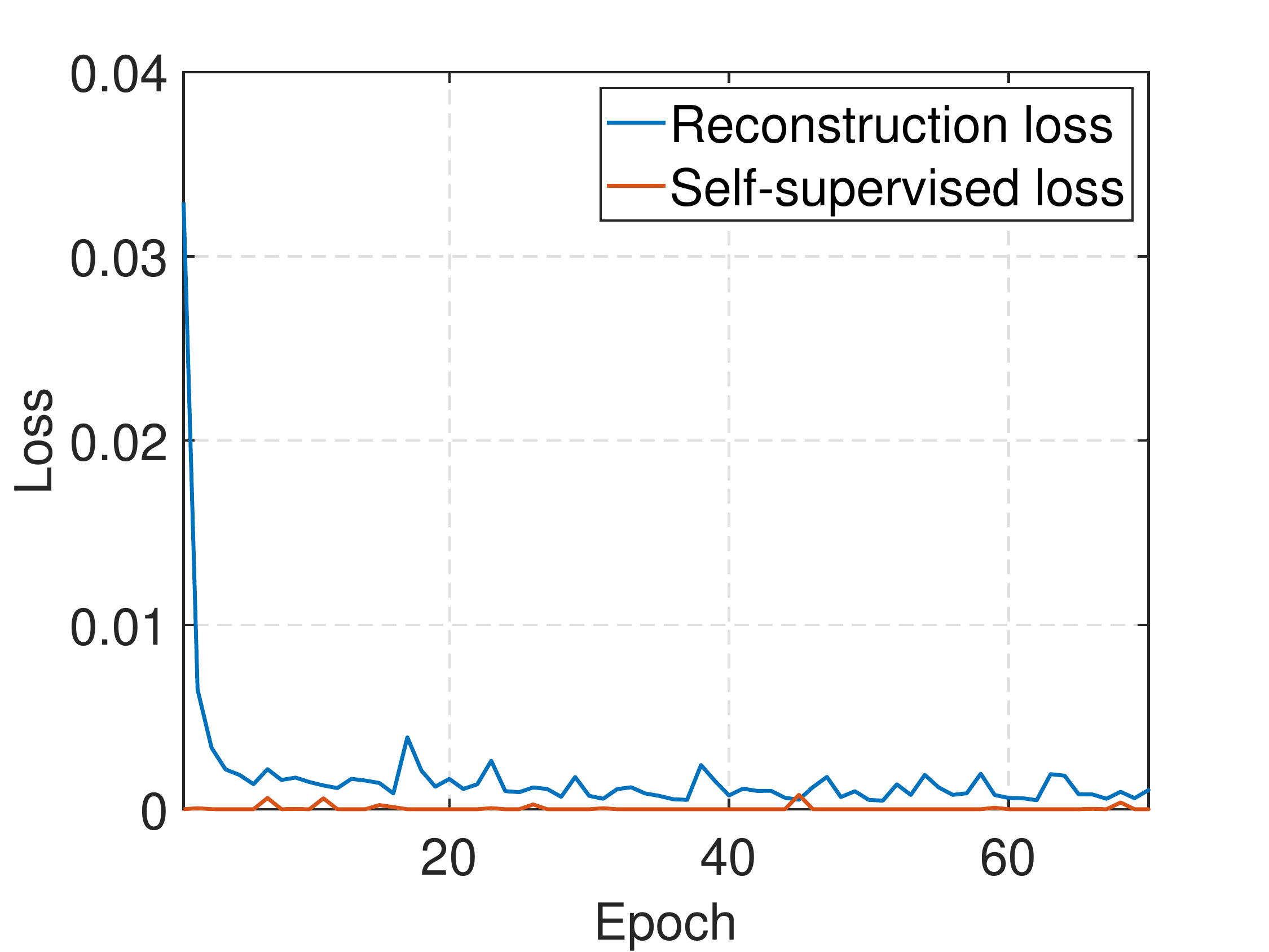}
        \label{fig-conver}
    }
    \subfigure[Inference time]{
        \includegraphics[width=.23\textwidth]{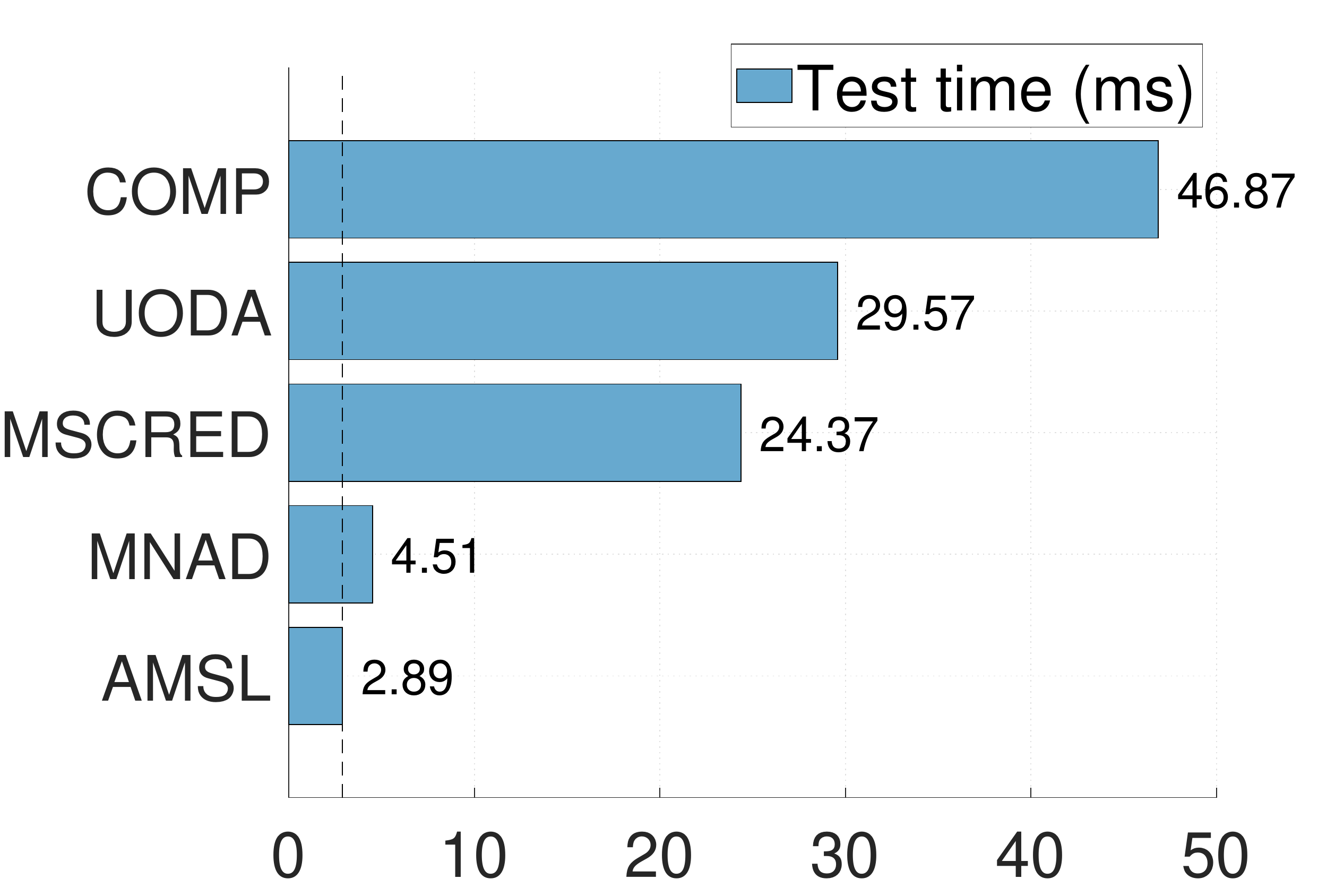}
        \label{fig-infer-time}
    }
    \caption{Convergence and inference time of \method.}
    \label{fig-1}
\end{figure}

\figurename~\ref{fig-conver} shows the convergence of reconstruction loss with memory modules along with the self-supervised loss. In conclusion, \method can be applied more effectively with a fast and stable convergence performance.

We also evaluate the inference time of \method and other strong baselines on DSADS dataset. Here, ``COMP'' means ConvLSTM-COMPOSITE. As shown in \figurename~\ref{fig-infer-time}, in addition to achieving the best performance, our method requires shorter running time than most other methods.

Besides, according to \tablename~\ref{tab-param} evaluated on DSADS dataset, the number of parameters and model size of \method are relatively smaller than most other methods. We also show that by reducing the model parameters through controlling the self-supervised data transformations $R$. We discard the poorly performing transformations in \tablename~\ref{fig-sub-ssl}. AMSL(R=6) discards the poorly performing transformation ``Noise'', AMSL(R=5) discards ``Noise'' and ``Scale'' transformations, AMSL(R=4) discards ``Noise'', ”Scale'' and ``Permuted” transformations, and AMSL(R=3) discards ``Noise'', ``Scale'', ``Permuted'' and ``Reversed'' transformations. We observe that our \method still achieves the best F1 and accuracy scores. On other datasets, the conclusions are also similar. This makes our method flexible in real applications.

\begin{table}[t!]
    \centering
    \caption{Comparison of model parameters.}
    \label{tab-param} 
    \begin{tabular}{ccccc}
    \toprule
        Method    & F1     & ACC    & \#Params & Model size \\ \hline
        MNAD      & 0.8003 & 0.7811 & 3.8M             & 14.7MB            \\
        MSCRED    & 0.6428 & 0.6192 & 3.6M             & 13.8MB            \\
        UODA      & 0.8475 & 0.8365 & 1.9M             & 7.3MB             \\ \hline
        AMSL(R=7) & 0.9352 & 0.9332 & 3.3M             & 12.7MB            \\
        AMSL(R=6) & 0.9298 & 0.9273 & 2.8M             & 10.9MB            \\
        AMSL(R=5) & 0.9242 & 0.9233 & 2.4M             & 9.2MB             \\
        AMSL(R=4) & 0.9138 & 0.9135 & 1.9M             & 7.4MB             \\
        AMSL(R=3) & 0.9099 & 0.9096 & 1.5M             & 5.7MB             \\     \bottomrule
    \end{tabular}
\end{table}

\section{Conclusions and Future Work}
\label{sec:con}
In this paper, we proposed an \methodfull (\method) for \uad on multivariate time series signals. To enhance the model's generalization ability towards unseen anomalies, we proposed to use the self-supervised learning module to learn diverse normal patterns and an adaptive memory fusion network to learn rich feature representations by the global and local memory modules. Experiments on four public datasets demonstrate that our methods significantly outperforms existing approaches in terms of accuracy, generalization, and robustness. 

In the future, we plan to extend \method to other modalities such as images and videos for \uad. In addition, we also plan to develop more efficient training algorithms and pursue the theoretical analysis of our method.

\section*{Acknowledgment}
This work is supported by the National Key Research and Development Plan of China (No. 2020YFC2007104), Natural Science Foundation of China (No. 61972383, No. 61902377, No. 61902379), Science and Technology Service Network Initiative, Chinese Academy of Sciences (No. KFJ-STS-QYZD-2021-11-001), the National Research Foundation, Singapore under its AI Singapore Programme (AISG Award No. AISG2-RP-2020-019), the RIE 2020 Advanced Manufacturing and Engineering (AME) Programmatic Fund (No. A20G8b0102), Singapore,  and the Nanyang Assistant Professorship (NAP).




\ifCLASSOPTIONcaptionsoff
  \newpage
\fi



%
\bibliographystyle{IEEEtran}
\bibliography{amsl_refs}



%






\end{document}